\documentclass{article}

\usepackage[preprint]{neurips_2026}

\usepackage[utf8]{inputenc}
\usepackage[T1]{fontenc}
\usepackage{hyperref}
\usepackage{url}
\usepackage{booktabs}
\usepackage{amsfonts}
\usepackage{amsmath}
\usepackage{amssymb}
\usepackage{amsthm}
\usepackage{graphicx}
\usepackage{microtype}
\usepackage[section]{placeins}
\usepackage{xcolor}
\usepackage{xspace}
\usepackage{enumitem}

\newcommand{\piz}{\ensuremath{\Pi_0}\xspace}
\newcommand{\pio}{\ensuremath{\Pi_1}\xspace}
\newcommand{\pit}{\ensuremath{\Pi_2}\xspace}
\newcommand{\pith}{\ensuremath{\Pi_3}\xspace}
\newcommand{\Dpart}{\ensuremath{\Delta_{\mathrm{partition}}}\xspace}

\newtheorem{lemma}{Lemma}
\newtheorem{theorem}{Theorem}
\newtheorem{corollary}{Corollary}
\newtheorem{proposition}{Proposition}
\theoremstyle{definition}
\newtheorem{definition}{Definition}
\title{A Regime Theory of Controller Class Selection for LLM Action Decisions}

\author{%
  Zhaoyang Jiang \\
  University of Glasgow \\
  \texttt{3167645J@student.gla.ac.uk} \\
  \And
  Zhizhong Fu \\
  UESTC, Chengdu, China \\
  \texttt{zhizhong.fu@std.uestc.edu.cn} \\
  \And
  Yunsoo Kim \\
  University College London \\
  \texttt{yunsoo.kim.23@ucl.ac.uk} \\
  \And
  Jiacong Mi \\
  University of Glasgow \\
  \texttt{j.mi.1@research.gla.ac.uk} \\
  \And
  Zicheng Li \\
  University of Glasgow \\
  \texttt{3222974L@student.gla.ac.uk} \\
  \And
  Xuanqi Peng \\
  University of Glasgow \\
  \texttt{3131960P@student.gla.ac.uk} \\
  \And
  Honghan Wu\thanks{Corresponding author.} \\
  University of Glasgow \\
  \texttt{Honghan.Wu@glasgow.ac.uk} \\
}

\begin{document}

\maketitle

\begin{abstract}
Deployed language and vision-language models must decide, on each input, whether to answer directly, retrieve evidence, defer to a stronger model, or abstain. Contrary to the common monotonicity intuition, greater per-input expressivity is not uniformly beneficial in finite samples: under identical strict cross-validation, different benchmarks prefer different controller classes. This reflects a finite-sample limitation of instance-level uncertainty signals, which can be exhausted at a distribution-dependent scale. We organize controllers into a nested lattice of four classes: fixed actions, partition routers, instance-level controllers, and prior-gated controllers, ordered by complexity. We prove a regime theory that turns three data-estimable bottlenecks into a class choice: how much improvement is possible beyond the best fixed action, whether there are enough samples for instance-level controllers to make reliable decisions, and how much improvement a coarse partition router can recover when instance-level signal is unreliable. The resulting Bernstein-tight threshold has a matching information-theoretic lower bound, and strict nested cross-validation provably selects a near-best class. Across SMS-Spam, HallusionBench, A-OKVQA, and FOLIO, the predicted class matches the empirical winner; the prior-gated controller wins on TextVQA when OCR tokens supply a label-free prediction-time prior. Code is available at \url{https://github.com/Anonymous-Awesome-Submissions/Regime-Theory}.
\end{abstract}

\section{Introduction}
\label{sec:intro}

Consider a medical question-answering system in clinical use. When asked ``what does this chest X-ray show?'', the system has several options: answer directly, retrieve relevant prior cases first, defer to a stronger specialist model, or abstain. A system that always answers, even when uncertain, is not just less accurate; it is dangerous. As large language and vision-language models move from benchmarks into deployment, \emph{when} to answer is becoming as consequential as how. A confidently wrong diagnosis, a fluent but factually incorrect claim, or a hard query handled by a weak model when a stronger one was available all share a common cause: the system chose the wrong action for an uncertain input.

The standard engineering response is to put a \emph{controller} in front of the model, with each subliterature settling on its own fixed structural design. The earliest version of the controller idea is the reject option: when confidence is too low, the system refuses to predict \citep{chow2003optimum, elyaniv2010selective, geifman2017selective}. The same logic reappears across modern generation systems as a decision to retrieve evidence before answering \citep{lewis2020rag, karpukhin2020dense, izacard2023atlas}, to pass a hard case to a stronger expert \citep{mozannar2020consistent, narasimhan2022posthoc, verma2023multiple, mao2023twostage}, or to climb a cost-ordered cascade of models \citep{chen2023frugalgpt, jitkrittum2023cascade, ding2024hybrid, ong2024routellm, dekoninck2024unified}. What changes across these settings is the action set and the surrogate; the underlying object is the same: a controller that maps each input to an action. Each subliterature then fixes, in advance, how expressive that map is allowed to be (for example, a global threshold, a calibrated selective score, an instance-level learned rule, or a cascade router), and optimizes within that fixed family, without asking whether the data support a coarser or finer controller class.

Our results contradict the implicit monotonicity premise in a structured way. Under an identical strict cross-validation protocol, the winning controller class changes with the statistical regime of the benchmark. On HallusionBench \citep{guan2024hallusionbench} and A-OKVQA \citep{schwenk2022aokvqa}, instance-level learned controllers win, consistent with strong per-sample signal and enough data to estimate it. On FOLIO \citep{han2024folio}, the sample size is far below the finite-sample viability threshold for instance-level control; a partition router reduces loss, while the best instance-level controller is slightly worse than the best fixed action. On SMS-Spam, the best fixed action already leaves very few inputs on which any alternative action can improve, and it remains the empirical winner (full numbers in Table~\ref{tab:main}). Thus the finite-sample ordering is not monotone in controller complexity: a coarser class can be the statistically justified choice even when a finer per-input rule is available. Two canonical cost-sensitive learning-to-defer baselines \citep{mozannar2020consistent, narasimhan2022posthoc}, both instance-level by construction, recover the high-signal gains on HallusionBench and A-OKVQA but fall behind the partition router on FOLIO. The effect is therefore about controller class, not a peculiarity of our implementation: the right question is not how to make the controller more expressive everywhere, but which class the data can actually support.

Two recent results suggest why. \citet{jitkrittum2023cascade} show that confidence-based cascade deferral is Bayes-optimal only when instance-level confidence is a sufficient statistic; \citet{kalai2024calibrated} and \citet{karbasi2025impossibility} prove impossibility-style lower bounds for hallucination detection from instance-level information alone. Both point to the same constraint: \emph{instance-level information is exhausted at a finite, distribution-dependent scale}. Neither, however, says how to choose a controller in practice. The question we therefore answer is: \emph{given the data, which controller class is statistically justified?}

\begin{figure}[t]
  \centering
  \includegraphics[width=0.95\linewidth]{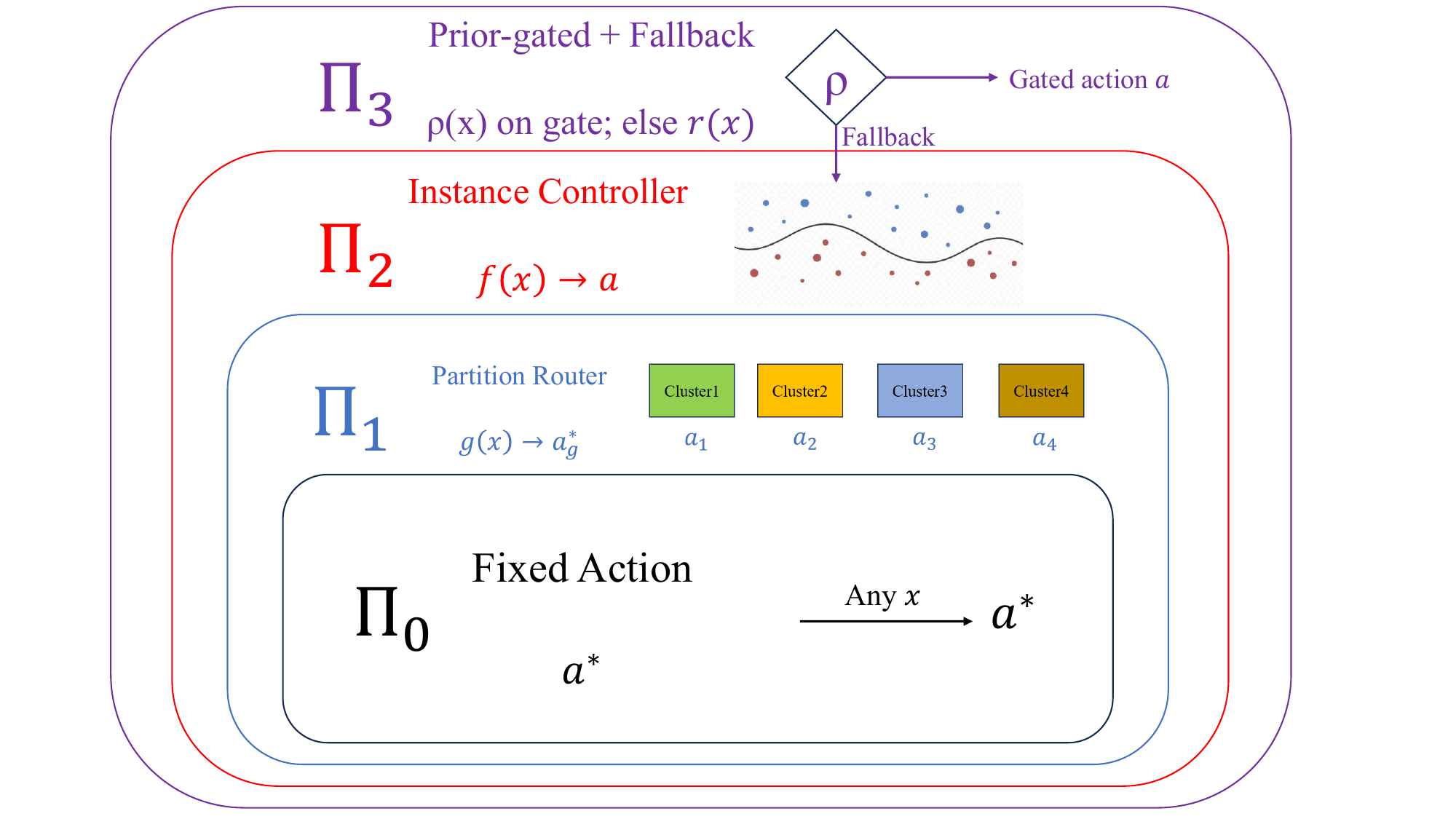}
  \caption{The nested lattice of policy classes. \piz contains fixed actions; \pio contains partition routers; \pit contains instance-level learned controllers; \pith pairs a deterministic prior gate with a fallback controller drawn from a lower class.}
  \label{fig:hierarchy}
\end{figure}

We organize controllers into a strictly nested four-class lattice \(\piz \subset \pio \subset \pit \subset \pith\) (Figure~\ref{fig:hierarchy}): fixed actions, partition routers that split inputs into a few groups, instance-level learned controllers, and prior-gated controllers that use an external signal when it is confident and otherwise fall back to a lower-class controller. The statistical point is simple: more complex controllers require enough samples before their gains can be certified, but passing this viability test does not by itself make a class the right choice. At a given \((\mathcal{D}, n)\) we should use the class whose certified gain best exceeds its estimation cost, which can be a lower-complexity class even when a higher one is statistically viable.

The theory makes this class choice operational. It reduces the question ``which controller class should be used at this data scale?'' to three data-estimable bottlenecks: whether there is enough residual mass for any adaptive controller to improve beyond the best fixed action, whether the available sample size can certify the sign of an instance-level \pit gain, and whether stable group-level action gaps allow a coarser \pio router to recover loss when per-sample control is unreliable. These bottlenecks are formalized by the residual bound, the Bernstein-tight \pit viability threshold with a matching information-theoretic lower bound, and the partition-gain bound (Theorems~\ref{thm:residual}--\ref{thm:partition}, Proposition~\ref{prop:lower}). Strict nested cross-validation then turns the diagnostics into a selection rule, choosing a near-best class up to an explicit stability term (Theorem~\ref{thm:oracle}). Empirically, the predicted class matches the empirical winner on all four core benchmarks; TextVQA-OCR provides the deployable \pith witness, with a $10.8$ joint seed-sd gap over the best lower class.

\section{Method}
\label{sec:method}

Section~\ref{sec:setup} defines the loss matrix and policy classes; Section~\ref{sec:theory} turns class choice into regime diagnostics and a nested-CV selector.

\subsection{Problem setup}
\label{sec:setup}

We consider an input \(x \sim \mathcal{D}\) and a finite action set \(\mathcal{A}=\{a_{\mathrm{direct}}, a_{\mathrm{retrieve}}, a_{\mathrm{defer}}, a_{\mathrm{abstain}}\}\). Each action \(a\) has correctness \(c(x,a) \in \{0,1\}\), semantic risk \(h(x,a) \in [0,1]\), and operational cost \(k(x,a) \ge 0\). The combined task loss is
\[
\ell(x,a) = w_{\mathrm{c}}(1-c(x,a)) + w_{\mathrm{h}} h(x,a) + w_{\mathrm{k}} k(x,a) \in [0, L_{\max}],
\]
with non-negative weights \((w_{\mathrm{c}}, w_{\mathrm{h}}, w_{\mathrm{k}})\) fixed by the deployment. A controller \(\pi: \mathcal{X} \to \mathcal{A}\) has population risk \(R(\pi) := \mathbb{E}_{\mathcal{D}}[\ell(x, \pi(x))]\). Setting \(w_{\mathrm{h}}=w_{\mathrm{k}}=0\) recovers raw zero-one accuracy with \(a_{\mathrm{abstain}}\) costing the same as a wrong direct prediction; classical reject-option theory \citep{chow2003optimum, franc2023optimal, pugnana2023auc} is the special case in which \(a_{\mathrm{abstain}}\) is assigned an intermediate cost \(c_r\in(0,1)\) via the semantic-risk channel (\(h(x,a_{\mathrm{abstain}})=c_r\), \(w_{\mathrm{h}}{=}1\)).

\begin{definition}[Policy classes]
\label{def:classes}
\piz is the class of \emph{fixed actions}, \(\pi_a(x) \equiv a\). \pio is the class of \emph{partition routers}: given a partition \(\{G_g\}_{g=1}^K\) and a per-cell action assignment \(a_g\), \(\pi(x) = a_g\) when \(x \in G_g\). \pit contains \pio plus \emph{instance-level learned controllers}: measurable maps from per-sample features through a bounded-complexity hypothesis family. \pith contains \pit plus \emph{prior-gated controllers}: \(\pi(x) = \rho(x)\) on inputs where an external prior channel is confident, and \(\pi(x) = r(x)\) otherwise, with \(r\) drawn from any lower class (\(r \in \piz \cup \pio \cup \pit\)). The deterministic rule \(\rho\) uses a channel not visible to \(r\)'s feature inputs.
\end{definition}

For the set inclusion, view \(x\) as the full prediction-time observation; lower classes simply ignore side channels. The four rungs are a constant action, a partition, a per-sample rule, and a side-information gate. If no label-free prediction-time prior is available, the deployable lattice truncates to \(\piz \cup \pio \cup \pit\).

The classes are nested as sets of functions, so population-optimal risk is monotone: \(R(\pi_0^\star) \ge R(\pi_1^\star) \ge R(\pi_2^\star) \ge R(\pi_3^\star)\). Estimation complexity goes the other way, and the question is which \(m\) minimizes the \emph{expected} risk \(\mathbb{E}_S[R(\widehat\pi_m(S))]\) at finite \(n\), and whether the answer can be identified from data.

\subsection{Finite-sample selection among policy classes}
\label{sec:theory}

We isolate three bottlenecks: residual mass, \pit sample size, and \pio partition geometry. Three theorems characterize them, one lemma and one lower bound support the \pit analysis, and a final theorem bounds held-out class selection. Throughout, let $a^\star := \arg\min_{a\in\mathcal{A}} \mathbb{E}[\ell(x,a)]$ denote the best fixed action and $R := \{x : \exists a, \ell(x,a) < \ell(x, a^\star)\}$ the residual set.

We first bound how much any adaptive controller can improve over the best fixed action. If only a small fraction of inputs admit a better action, then the total gain is limited by that fraction times the largest possible per-input improvement.

\begin{theorem}[Residual bound]
\label{thm:residual}
For any controller $\pi: \mathcal{X} \to \mathcal{A}$,
\[
\mathbb{E}[\ell(x, a^\star) - \ell(x, \pi(x))] \;\le\; \mathbb{P}(R) \cdot \sup_{x \in R, a \in \mathcal{A}}(\ell(x, a^\star) - \ell(x, a)).
\]
\end{theorem}

The proof (Appendix~\ref{app:proofs}) is a one-line conditioning argument. We next analyze a canonical selective subproblem inside \pit only to derive the finite-sample threshold; the evaluated \pit families in Section~\ref{sec:experiments} remain multi-action.

A score \(f\) induces a rejector \(\pi_q\): it plays \(a_{\mathrm{direct}}\) on the top \(1{-}q\) fraction of inputs, ranked by \(f\), and a fallback action on the bottom \(q\). Let \(L_r, L_w, L_a\) be the expected losses of \(a_{\mathrm{direct}}\) on correct inputs, \(a_{\mathrm{direct}}\) on wrong inputs, and the fallback. Define
\[
\alpha_{\min} := \tfrac{L_a - L_r}{L_w - L_r}, \qquad \beta := \alpha_{\mathrm{emp}} - \alpha_{\min},
\]
where $\alpha(f)$ is the AUC of $f$ against direct correctness and $\alpha_{\mathrm{emp}} = \sup_f \alpha(f)$ is the AUC ceiling on the feature class. Thus $\alpha_{\min}$ is the break-even AUC for routing the bottom-$q$ to a fallback, and $\beta = \alpha - \alpha_{\min}$ is the margin above break-even.

The asymptotic improvement condition is classical \citep{chow2003optimum, franc2023optimal, pugnana2023auc}: under the local tail-margin convention introduced in Lemma~\ref{lem:necessary}, $\pi_q$ improves on $a_{\mathrm{direct}}$ asymptotically when $\beta > 0$, with reduction of order $q^\star \beta (L_w - L_r)$ at optimal $q^\star$ (the exact iff condition is on $\mu_w(q)-\alpha_{\min}$, for which $\beta$ is the AUC-level scalar proxy).

\begin{lemma}[Asymptotic signal condition, after \citep{chow2003optimum, franc2023optimal}]
\label{lem:necessary}
The exact improvement condition is on the bottom-\(q\) tail precision: \(\pi_q\) improves on \(a_{\mathrm{direct}}\) asymptotically iff \(\mu_w(q)>\alpha_{\min}\) for some \(q\). Suppose \(\mu_w(q)-\alpha_{\min}\) shares sign with the AUC margin \(\beta=\alpha_{\mathrm{emp}}-\alpha_{\min}\) at the operating \(q\). We use this structural condition throughout as the local tail-margin convention. Under this convention, \(\beta\) is a sufficient scalar proxy: \(\beta\le0\) rules out asymptotic improvement, and \(\beta>0\) admits scale \(q^\star\beta(L_w-L_r)\) at optimal coverage \(q^\star\).
\end{lemma}

The new finite-sample result quantifies how large $n$ must be for the population gain to be visible above estimator variance.

\begin{theorem}[Bernstein-tight viability threshold]
\label{thm:bernstein}
Let $\ell \in [0, L_{\max}]$, $\alpha \ge 1/2$, and $\beta = \alpha - \alpha_{\min} > 0$. Assume the local tail-margin condition $\mu_w(q) - \alpha_{\min} \ge \beta$, where $\mu_w(q) := \mathbb{P}(c_{\mathrm{direct}}{=}0 \mid f(x) \in \mathrm{bottom}\text{-}q)$. Then to leading order in $n$,
\begin{equation}
\label{eq:bern}
n \;\ge\; n_{\min}(\alpha, \beta, q, \delta) \;:=\; \frac{2\alpha(1-\alpha)\log(2/\delta)}{q\beta^2}
\end{equation}
is sufficient for the empirical CV loss difference $\widehat{\Delta}(q)$ to have the same sign as the population loss difference with probability at least $1 - \delta$. When $n$ is materially below~\eqref{eq:bern}, the bound no longer certifies sign correctness; we call this regime \emph{variance-bounded}.
\end{theorem}

\begin{figure}[t]
  \centering
  \includegraphics[width=0.62\linewidth]{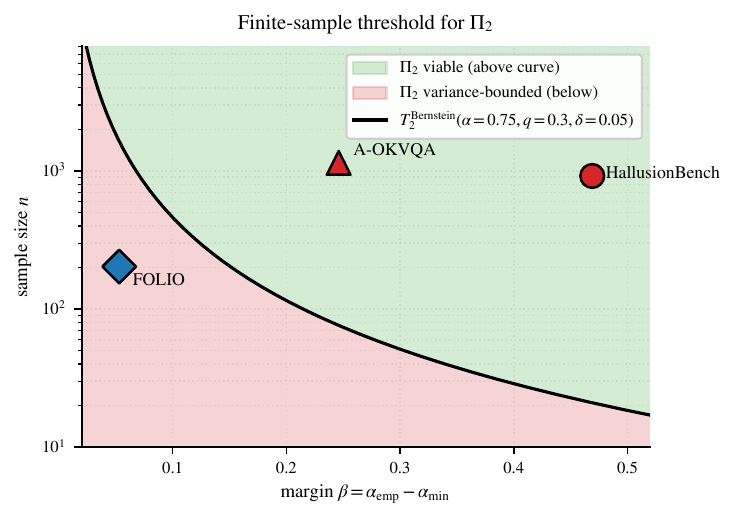}
  \caption{The viability threshold from~\eqref{eq:bern} at $\alpha{=}0.75, q{=}0.3, \delta{=}0.05$. Green: \pit-viable. Red: variance-bounded. FOLIO is the one core benchmark far below the Bernstein threshold (\(n{=}203\) vs.\ \(n_{\min}{=}1898\)) and lands deep in the variance-bounded region; its best empirical \pit controller slightly increases loss relative to \piz (\(+0.003\)), consistent with the uncertified-sign regime of Corollary~\ref{cor:nested}~(ii). HallusionBench (\(n{=}920\gg n_{\min}{=}23\)) and A-OKVQA (\(n{=}1145\gg n_{\min}{=}45\)) sit comfortably above their respective thresholds. SMS-Spam is governed instead by the residual bound of Theorem~\ref{thm:residual}: its \(\beta\le0\) violates the positive-margin condition of Theorem~\ref{thm:bernstein}, so it lies outside the theorem's viability-threshold regime.}
  \label{fig:phase}
\end{figure}

The threshold is computable from $(\alpha, \beta, q, \delta)$ alone (Figure~\ref{fig:phase}). Theorem~\ref{thm:bernstein} is the finite-sample counterpart of \citet{jitkrittum2023cascade}: even with positive asymptotic margin, the sign of a controller's empirical gain can be uncertified when $n$ is too small. The proof (Appendix~\ref{app:bernstein}) applies Bernstein's inequality to the bottom-$q$ precision estimator. The next proposition shows that the same dependence on $\alpha$, $\beta$, $q$, and $\delta$ is necessary up to constants.

\begin{proposition}[Information-theoretic lower bound]
\label{prop:lower}
In the setting of Theorem~\ref{thm:bernstein}, suppose the bottom-\(q\) decision reduces locally to testing \(H_0:\mu_w=\alpha_{\min}\) against \(H_1:\mu_w=\alpha_{\min}+\beta\), with \(0<\beta\le \tfrac12\min\{\alpha_{\min},1-\alpha_{\min}\}\). Any test that determines the sign of the population loss difference \(\Delta(q)\) with both type-I and type-II error at most \(\delta\in(0,\tfrac18)\) from \(n\) i.i.d.\ samples must satisfy
\begin{equation}
\label{eq:lower}
n \;\ge\; \frac{c\,\alpha(1{-}\alpha)\,\log(1/\delta)}{q\,\beta^{2}}
\end{equation}
for a universal constant \(c>0\). Since the sufficient condition~\eqref{eq:bern} has the same functional form \(n_{\min}=2\alpha(1{-}\alpha)\log(2/\delta)/(q\beta^{2})\), the viability threshold has the optimal high-confidence order under this reduction: \(n_{\min}\asymp\alpha(1{-}\alpha)\log(1/\delta)/(q\beta^{2})\), up to universal constants and the local replacement of \(\alpha_{\min}\) by \(\alpha=\alpha_{\min}+\beta\).
\end{proposition}

The proof (Appendix~\ref{app:lower}) is Bretagnolle--Huber on the local Bernoulli alternative \citep{tsybakov2009nonparametric}. Thus, under the local testing reduction, the threshold is order-tight.

A coarser controller can win where \pit is variance-bounded, not because it is more powerful, but because it asks less of the data: a partition router only needs each group to have a clear best action, not a precise per-sample score, so its sample complexity depends on cluster mass and gap rather than on the per-sample margin that controls \pit. Let $\mathcal{G} = \{G_g\}_{g=1}^K$ be a measurable partition with $p_g := \mathbb{P}(x \in G_g)$, cell-best actions $a_g^\star := \arg\min_a \mathbb{E}[\ell(x,a) \mid x \in G_g]$, and discriminability gaps $\gamma_g := \mathbb{E}[\ell(x, a^\star) - \ell(x, a_g^\star) \mid x \in G_g] \ge 0$.

\begin{theorem}[Distribution-dependent partition lower bound]
\label{thm:partition}
For any measurable partition \(\mathcal{G}\) with cluster probabilities \(p_g\) and discriminability gaps \(\gamma_g\),
\begin{equation}
\label{eq:partition}
\Dpart(\mathcal{G})\;=\;\sum_{g=1}^{K}p_g\gamma_g\;\ge\;\max_{g}\,p_g\gamma_g.
\end{equation}
Moreover, let \(\sigma^{2}\) be an upper bound on the per-action conditional loss variance on any cell. Under Bernstein concentration on each cell, the empirical \pio estimator with \(\kappa\)-fold CV on \(n\) samples achieves cell-loss no worse than the global baseline \(a^\star\) on cluster \(g\) (i.e.\ secures the corresponding \(p_g\gamma_g\) contribution to~\eqref{eq:partition}) with probability at least \(1-\delta\) whenever
\(
n \;\ge\; \tfrac{\kappa}{\kappa-1}\max\!\left\{8\log(2/\delta)/p_g,\;(16\sigma^{2}+(8/3)L_{\max}\gamma_g)\log(4|\mathcal{A}|/\delta)/(p_g\gamma_g^{2})\right\},
\)
where the first term is a Chernoff occupancy floor and the second the Bernstein concentration term. The sample complexity depends on \(p_g\) and \(\gamma_g\) but not on the instance-level margin \(\beta\).
\end{theorem}

The sample complexity scales with $p_g$ and $\gamma_g$, \emph{not} with the per-sample margin $\beta$ that controls \pit, so a partition router can succeed precisely where an instance-level controller is variance-bounded. The $\max_g p_g \gamma_g$ form is also diagnostic: when the empirical reduction is close to it, the partition win is essentially a one-cluster phenomenon. Proof is in Appendix~\ref{app:proofs}; the bound is agnostic, and an adaptive partition such as KMeans can be tighter, which we treat as an empirical observation (Appendix~\ref{app:cluster}).

For comparing supportable adaptive classes, write
\[
C_{\Pi_1}(\mathcal{G}) := \sum_g p_g\gamma_g, \qquad
C_{\Pi_1} := \max_{\mathcal{G}\in\mathfrak{G}} C_{\Pi_1}(\mathcal{G}),
\]
for a candidate partition family \(\mathfrak{G}\), and
\[
C_{\Pi_2} := q^\star\beta(L_w-L_r)
\]
for the AUC-level \pit ceiling. The residual product, the \pit viability test ($n \gtrless n_{\min}$), and the two ceilings \(C_{\Pi_1}, C_{\Pi_2}\) give the regime decomposition:

\begin{corollary}[Regime decomposition]
\label{cor:nested}
For $(\mathcal{D}, n)$, (i) every adaptive class is capped by the residual product in Theorem~\ref{thm:residual}; when this product is small, no adaptive class can improve by more than residual scale. (ii) If $\beta > 0$ but $n < n_{\min}$, $\pit$'s sign is uncertified, while $\pio$ can strictly improve on $\piz$ whenever some candidate partition has \(C_{\Pi_1}(\mathcal{G})>0\). (iii) If $n \ge n_{\min}$, the population-optimal winner between $\pio$ and $\pit$ is determined by comparison between the ceilings \(C_{\Pi_1}\) and \(C_{\Pi_2}\); the empirical winner additionally depends on how tightly the deployed candidate families realize each ceiling at finite \(n\).
\end{corollary}

When benchmarks fall into different cases, no single controller family is uniformly optimal. Corollary~\ref{cor:nested} is the regime diagnostic; Section~\ref{sec:experiments} uses strict nested CV over candidate classes, and Theorem~\ref{thm:oracle} bounds the held-out selection cost. The bound is stated for the inner-CV estimand because validation-based selection can itself overfit \citep{cawley2010overfitting}.

\begin{theorem}[Held-out selection bound]
\label{thm:oracle}
Let $\mathcal{M} \subseteq \{0,1,2,3\}$ be the candidate classes ($|\mathcal{M}|=3$ when no deployable external channel is available). Let $\widehat R_{\mathrm{CV}}(m)$ be the inner $\kappa$-fold CV estimate and $\bar R_\kappa(m)$ the expected risk of an inner-trained estimator. With $\widehat m := \arg\min_{m \in \mathcal{M}} \widehat R_{\mathrm{CV}}(m)$ and $\ell \in [0, L_{\max}]$,
\begin{equation}
\label{eq:oracle}
\mathbb{E}[\bar R_\kappa(\widehat m)] \;\le\; \min_{m \in \mathcal{M}} \bar R_\kappa(m) + 4 L_{\max} \sqrt{\log(2|\mathcal{M}|)/n_{\mathrm{in}}},
\end{equation}
where $n_{\mathrm{in}} = n_{\mathrm{out}}/\kappa$. If each estimator is $\varepsilon_{\mathrm{stab}}$-stable when refit from inner-train to outer-train, the bound transfers to the refit estimator with an added $2\varepsilon_{\mathrm{stab}}$ term.
\end{theorem}

The proof (Appendix~\ref{app:oracle}) is Hoeffding plus a union bound over $\mathcal{M}$. The $\min_m$ on the right absorbs the class-specific estimation errors controlled by Theorems~\ref{thm:bernstein}--\ref{thm:partition}: zero for $m{=}0$, partition complexity for $m{=}1$, and the Bernstein threshold for $m{=}2$.

\section{Experiments}
\label{sec:experiments}
\label{sec:arc}

\paragraph{Experimental protocol and data.} We instantiate the lattice as a controller pool under strict nested 5-fold-by-5-seed CV: family selection on outer-train via inner CV, refit on outer-train, single evaluation on outer-test. The pool keeps \piz at always-direct and a fair-fixed CV variant; \pio uses KMeans partition routers (\(K\in\{4,5,6,8\}\)); \pit uses HGBC (\(\mathrm{md}\in\{3,4\}\)) and a calibrated logistic plug-in (\(C{=}0.3\)). Hyperparameters are fixed in advance, not tuned per benchmark; a structurally different CART partition variant of \pio is reported as an ablation (Appendix~\ref{app:ablations}). Reported \(\pm\) values are 5-seed standard deviations; gaps in seed-sd units reflect algorithmic stability rather than dataset-resampling error. Implementation details are in Appendices~\ref{app:l2d}--\ref{app:protocol}; cost-sensitive L2D adaptations \citep{mozannar2020consistent, narasimhan2022posthoc} are reported in Appendix~\ref{app:l2d-results}.

We report results on four full-corpus benchmarks; each regime tag follows from \((\alpha, \beta, n, \{p_g\gamma_g\})\) computed on the held-out split, independent of any per-class CV outcome. The benchmarks: SMS-Spam \citep{almeida2011smsspam} (saturated text, residual-bounded), HallusionBench \citep{guan2024hallusionbench} (visual hallucination, large-\(\beta\)), A-OKVQA \citep{schwenk2022aokvqa} (near-tie \pio/\pit), and FOLIO \citep{han2024folio} (small \(\beta\) and \(n{<}n_{\min}^{\mathrm{Bern}}\), variance-bounded). On HallusionBench/A-OKVQA, \(a_{\mathrm{direct}}\) and \(a_{\mathrm{retrieve}}\) are produced by Qwen2.5-VL-3B-Instruct (\(a_{\mathrm{retrieve}}\): 2-turn grounded-context pass) and \(a_{\mathrm{defer}}\) by Qwen2.5-VL-7B-Instruct (\(c_{\mathrm{defer}}/c_{\mathrm{direct}}{=}2.3\)); FOLIO uses the same backbones in text-only mode. Per-action risk \(h(x,a)\in\{0,0.5,1\}\) on HallusionBench/A-OKVQA comes from an InternVL2.5-8B \citep{chen2024internvl} judge \emph{drawn from a different VLM family than the upstream Qwen2.5-VL policies}, so the judge does not score outputs from its own family; FOLIO uses a rule-based risk channel (\(h{=}0.5\) on wrong, \(0\) on correct), and SMS-Spam uses canonical-cost weighting without a judge. Canonical weights \((w_{\mathrm{c}}, w_{\mathrm{h}}, w_{\mathrm{k}})=(1,1,0.05)\); weight robustness in Appendix~\ref{app:ablations}.

Controllers consume a scalar feature block of 41 dims on HallusionBench, 39 dims on A-OKVQA, 12 dims on FOLIO/SMS-Spam, all derived from upstream Qwen2.5-VL outputs at prediction time. The InternVL2.5-8B judge is used only to label semantic risk in the loss matrix and is never exposed to the controller (Appendices~\ref{app:per-bench},~\ref{app:loss}). None of the four core benchmarks supplies a label-free prediction-time prior channel, so the deployable lattice truncates to \(\piz \cup \pio \cup \pit\); we use TextVQA \citep{singh2019textvqa} for a separate \pith{} test.

Each benchmark uses its full native held-out split (\(n\) values in Table~\ref{tab:main}; HallusionBench drops \(31\) of \(951\) rows with empty Qwen-VL-3B commitments), spanning four \((\alpha,\beta,n,\{p_g\gamma_g\})\) regimes; the controlled \(n\)-sweep is deferred to Appendix~\ref{app:bernstein-cross}.

\begin{table}[t]
  \centering
  \small
  \resizebox{\linewidth}{!}{
  \begin{tabular}{lcccc}
    \toprule
    Benchmark & \piz best$\downarrow$ & \pio best$\downarrow$ & \pit best$\downarrow$ & predicted \\
    \midrule
    SMS-Spam (\(n{=}1114\))      & \(\mathbf{0.0590\pm0.0000}\) & \(\mathbf{0.0590\pm0.0000}\)\,$^{\dagger}$ & \(0.0591\pm0.0000\) & \piz\\
    HallusionBench (\(n{=}920\)) & \(1.0000\pm0.0000\) & \(0.9093\pm0.0025\) & \(\mathbf{0.8970\pm0.0039}\) & \pit\\
    A-OKVQA (\(n{=}1145\))       & \(0.4138\pm0.0000\) & \(0.3902\pm0.0085\) & \(\mathbf{0.3805\pm0.0032}\) & \pit\\
    FOLIO (\(n{=}203\))          & \(0.7520\pm0.0000\) & \(\mathbf{0.7195\pm0.0073}\) & \(0.7546\pm0.0086\) & \pio\\
    \bottomrule
  \end{tabular}}
  \caption{Strict nested 5-fold-by-5-seed CV per-class loss on the four core benchmarks (lower is better; values are mean \(\pm\) seed-sd, fixed policies have zero seed variance). Bold marks the empirical winner within \(\piz\cup\pio\cup\pit\); the rightmost column gives the regime-theoretic predicted winner from Theorem~\ref{thm:bernstein} and Corollary~\ref{cor:nested} (full predictions with diagnostics in Table~\ref{tab:theory}). \(\dagger\): on SMS-Spam, \pio collapses to \(a_{\mathrm{direct}}\) under CV, so \(\pio = \piz\) numerically; \piz is the strict winner because \pit is \(+0.0002\) worse. Per-benchmark mechanism (including \texttt{fair\_fixed\_train} on HallusionBench and the uncertified \pit outcome on FOLIO) is unpacked in the body.}
  \label{tab:main}
\end{table}

\begin{figure}[t]
  \centering
  \includegraphics[width=\linewidth]{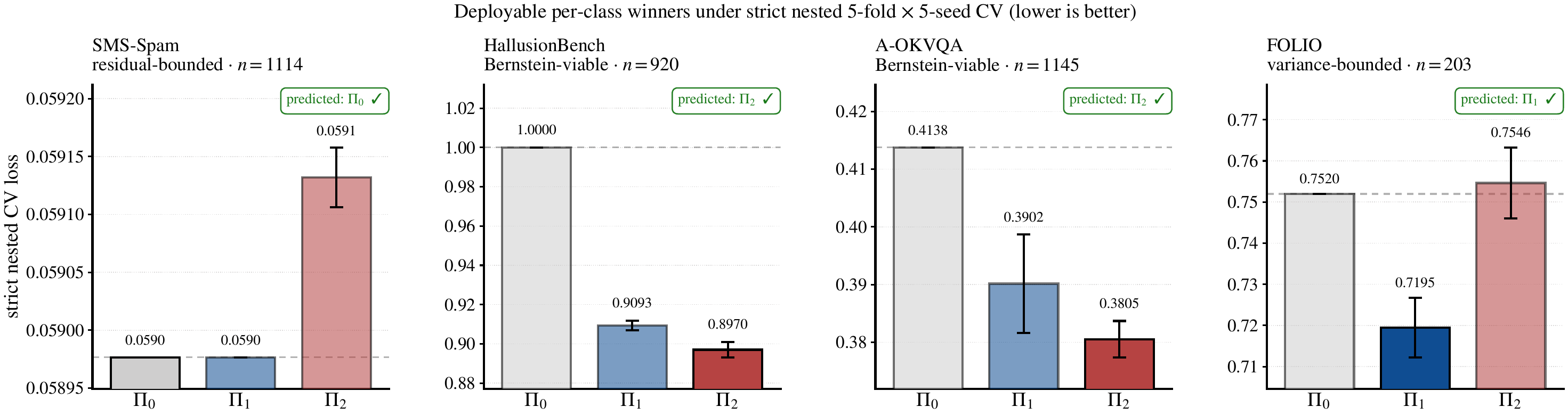}
  \caption{Deployable per-class winners under strict nested 5-fold-by-5-seed CV on the four core benchmarks (lower is better). Error bars are multi-seed standard deviations. Gold outlines mark the deployable per-benchmark winner within \(\piz\cup\pio\cup\pit\). The winning class changes across benchmarks in the pattern indicated by the regime diagnostics: \piz on SMS-Spam (residual-bounded), \pit on HallusionBench and A-OKVQA (Bernstein-viable, with A-OKVQA near the \pio/\pit boundary), and \pio on FOLIO (variance-bounded; \pit's sign uncertified). The non-deployable A-OKVQA \pith gold-rationale oracle is reported separately in Table~\ref{tab:pi3-real}.}
  \label{fig:per-class}
\end{figure}

\paragraph{Main result.} Table~\ref{tab:main} and Figure~\ref{fig:per-class} give deployable per-class winners; the pattern aligns with Corollary~\ref{cor:nested}. \emph{SMS-Spam} (residual-bounded, $\mathbb{P}(R){=}0.009$): every \pio family collapses to direct in CV and the best \pit family is $+0.0002$ worse, so \piz strictly wins (Appendix~\ref{app:sms-spam}). \emph{HallusionBench} (Bernstein-viable, $n{=}920\gg n_{\min}^{\mathrm{Bern}}{=}23$): \pit (HGBC-\texttt{md3}, $-0.103$, $26$ seed-sd) edges \pio (KMeans-$K{=}8$, $-0.091$) by $\approx 3$ seed-sd; inner-CV picks \pit on $20/25$ outer-folds (Appendix~\ref{app:auto-select}); the instance ceiling exceeds the partition ceiling (Cor~\ref{cor:nested}~(iii)). \emph{A-OKVQA} (\pio/\pit boundary): \pit (Selective-$C{=}0.3$, $-0.033$) edges \pio (KMeans-$K{=}4$, $-0.024$) by $\approx 1$ seed-sd; inner-CV splits \pio:\pit as $10{:}15$ over $25$ folds. \emph{FOLIO} (variance-bounded, $n{=}203 \ll n_{\min}^{\mathrm{Bern}}{=}1898$; Cor~\ref{cor:nested}~(ii)): \pio (KMeans-$K{=}6$, $-0.033$, $4.5$ seed-sd) wins; the best \pit controller slightly increases loss relative to \piz ($+0.0027$), and cost-sensitive L2D adaptations \citep{mozannar2020consistent, narasimhan2022posthoc} (both \pit-class) reach $0.7478$ and $0.7705$, trailing our \pio by $0.028$/$0.051$ (Appendix~\ref{app:l2d-results}). The A-OKVQA \(\pith\) rationale experiment is excluded from the deployable comparison because rationales are answer-derived (Appendix~\ref{app:pi3-rationale}).

\begin{table}[t]
  \centering
  \small
  \resizebox{\linewidth}{!}{
  \begin{tabular}{lcccccccc}
    \toprule
    Benchmark & \(n\) & \(\alpha_{\mathrm{emp}}\) & \(\beta\) & \(n\beta^{2}\) & \(n_{\min}^{\mathrm{Bern}}\) & \(C_{\Pi_1}\) & \(C_{\Pi_2}\) & predicted class\\
    \midrule
    SMS-Spam       & 1114 & 0.991 & \(\le 0\)\textsuperscript{$\dagger$} & --- & --- & \(0.000\) & --- & \piz (residual-bounded; Thm~\ref{thm:residual}) \\
    HallusionBench & 920  & 0.722 & 0.469 & 202.4 & 23   & \(0.10\)  & \(\mathbf{0.15}\) & \pit (high-signal, $n{\gg}n_{\min}$; Thm~\ref{thm:bernstein}) \\
    A-OKVQA        & 1145 & 0.874 & 0.246 & 69.3  & 45   & \(0.039\) & \(\mathbf{0.094}\) & \pit (high-signal, \pio/\pit boundary) \\
    FOLIO          & 203  & 0.687 & 0.053 & 0.57  & 1898 & \(\mathbf{0.065}\) & \(0.024\)\textsuperscript{$\ddagger$} & \pio ($n{<}n_{\min}$, \pit uncertified; Cor~\ref{cor:nested} (ii)) \\
    \bottomrule
  \end{tabular}}
  \caption{Theory-instantiated quantities from Theorem~\ref{thm:bernstein} and Corollary~\ref{cor:nested}. \(n_{\min}^{\mathrm{Bern}}\) is computed from the leading-order formula \(\lceil 2\alpha(1{-}\alpha)\log(2/\delta)/(q\beta^{2})\rceil\) at \(\delta{=}0.05\), \(q{=}0.3\), using the cross-validated AUC margin as the local tail-margin proxy. \(C_{\Pi_1}=\max_{\mathcal{G}\in\mathfrak{G}}\sum_g p_g\gamma_g\) over the candidate KMeans family; \(C_{\Pi_2}=q^\star\beta(L_w-L_r)\). Bold marks the diagnostic favoring the predicted adaptive class: for HallusionBench and A-OKVQA both classes are supportable and the larger ceiling indicates the Cor~\ref{cor:nested}~(iii) winner; for FOLIO \pit is uncertified ($n{<}n_{\min}^{\mathrm{Bern}}$, Cor~\ref{cor:nested}~(ii)) so $C_{\Pi_1}$ is bolded as the supportable winner. \textsuperscript{$\dagger$}SMS-Spam: \(\beta\le 0\), Lemma~\ref{lem:necessary} rules out an instance-level rejector; \(\mathbb{P}(R){=}0.009\), oracle \(\Pi_2\) gain \({=}0.003\). \textsuperscript{$\ddagger$}FOLIO: \(C_{\Pi_2}\) is the asymptotic ceiling but \(\pit\)'s sign is uncertified at \(n{=}203<n_{\min}^{\mathrm{Bern}}\), so the empirical winner falls to \pio. Diagnostics are reported post-hoc on outer-test concatenations; the strict-CV selection rule uses only inner-fold quantities (Theorem~\ref{thm:oracle}; auto-pick analysis in Appendix~\ref{app:auto-select}).}
  \label{tab:theory}
\end{table}

\paragraph{Theory-to-benchmark match.}\label{sec:theory-match}
Table~\ref{tab:theory} instantiates the diagnostics behind Figure~\ref{fig:per-class}. FOLIO lies far below the Bernstein threshold ($n\beta^2{=}0.57$, $n_{\min}{=}1898$), where \pit's sign is uncertified (Cor~\ref{cor:nested}~(ii)) and the supportable adaptive winner is \pio. HallusionBench resolves via $C_{\Pi_2}{>}C_{\Pi_1}$ (Cor~\ref{cor:nested}~(iii), \pit winner); A-OKVQA is a \pio/\pit boundary case, reflected in the $10{:}15$ inner-CV class split (Appendix~\ref{app:auto-select}); SMS-Spam has $\beta\le0$ and is governed by the residual bound (Thm~\ref{thm:residual}; Appendix~\ref{app:sms-spam}). The classification is unchanged at $q\in\{0.2,0.3,0.4\}$.

\paragraph{TextVQA-OCR: deployable \pith.}\label{sec:pi3-textvqa} The four core benchmarks above span the deployable \(\piz \cup \pio \cup \pit\) lattice; we now turn to the top rung. \pith{} requires a label-free prediction-time prior channel, and TextVQA is the one benchmark in our pool that supplies one: OCR tokens are image-derived and observable before answering. We freeze two upstream actions on the official training split (a question/image-class direct predictor and an OCR-copy predictor), then route on the official validation split ($n{=}5000$) under strict 5-fold-by-5-seed CV. Lower rungs see only scalar non-OCR features (the \pio pool here also includes a CART variant over that block). \pith alone gates on OCR-prior confidence and sends the rest to a learned fallback. Loss is $1{-}$soft-VQA accuracy.

\begin{table}[t]
  \centering
  \small
  \begin{tabular}{lccc}
    \toprule
    TextVQA-OCR validation ($n{=}5000$) & best family & answer loss & $\Delta$ vs \pith \\
    \midrule
    \piz & \texttt{always\_ocr} & $0.8463\pm0.0000$ & $+0.0251$\\
    \pio & CART-\texttt{d4} & $0.8445\pm0.0006$ & $+0.0233$\\
    \pit & logistic $C{=}0.3$ & $0.8595\pm0.0021$ & $+0.0383$\\
    \pith & OCR gate + fallback & $\mathbf{0.8212\pm0.0021}$ & $\mathbf{0}$\\
    \bottomrule
  \end{tabular}
  \caption{Deployable \pith on TextVQA-OCR. Lower rungs see only scalar non-OCR features; \pith alone gates on OCR-prior confidence. \pit underperforms \piz because non-OCR features cannot recover OCR-side information (information-design issue, not a Bernstein-threshold failure: $n{=}5000$ is well above $n_{\min}$). \pith's $0.0233$ gap over best-lower \pio is $10.8$ joint seed-sd ($\sigma_{\mathrm{j}}{=}0.00217$).}
  \label{tab:pi3-textvqa}
\end{table}

The fixed OCR action reaches $15.4\%$ soft accuracy versus direct's $8.4\%$, so the benchmark contains usable side information that the scalar feature block cannot recover. \pit on non-OCR features is actually \emph{worse} than the fixed OCR action ($0.8595$ vs.\ $0.8463$): an information-design issue, not a Bernstein-threshold failure at $n{=}5000$. \pith resolves this by gating high-confidence OCR cases on the prior and sending the rest to a learned fallback, reducing loss by $0.0233$ over the best lower class. The synthetic in Appendix~\ref{app:pi3-synth} constructs the prior independent of \(X\), confirming \pit's lack of access cannot be rescued by additional samples.

\paragraph{Additional checks.} Appendix analyses corroborate the regime map: a bottom-$q$ precision-estimator cross-threshold synthetic for Theorem~\ref{thm:bernstein} that sweeps $n$ across $n_{\min}(\beta)$ and recovers the predicted sign-correctness transition (Appendix~\ref{app:bernstein-cross}); the \pio/\pit and \pith synthetics (Appendices~\ref{app:synth-pi12},~\ref{app:pi3-synth}); HallusionBench cluster anatomy with $\sum_g p_g\gamma_g{=}0.047$ at $K{=}4$ saturating the empirical KM-$K{=}4$ loss reduction $0.048\pm0.005$ (Appendix~\ref{app:cluster}); CART and loss-weight ablations (Appendix~\ref{app:ablations}).

\section{Related Work}
\label{sec:related}

\paragraph{Selective classification, learning to defer, and cascade routing.} A long line of work fixes a single controller family and refines the score, surrogate, or calibrator within it---from classical reject-option theory \citep{chow2003optimum, elyaniv2010selective, geifman2017selective, franc2023optimal, pugnana2023auc, traub2024overcoming} to cost-sensitive deferral \citep{mozannar2020consistent, narasimhan2022posthoc, verma2023multiple, mao2023twostage}. Cascade routing \citep{chen2023frugalgpt, ding2024hybrid, ong2024routellm, chen2024routerdc, dekoninck2024unified} extends this to a cost-ordered model sequence but keeps the family fixed; \citet{goren2024hierarchical} nest abstention in the \emph{output} space rather than in the controller space we study. Conformal prediction \citep{angelopoulos2023conformal, bates2021distribution} calibrates a threshold inside a fixed selective family. We instead ask which family the available $n$ can support: our viability threshold is the finite-sample counterpart of \citet{jitkrittum2023cascade}'s sufficient-statistic characterization, and our held-out selection bound (Theorem~\ref{thm:oracle}) adapts classical nested-CV concerns \citep{cawley2010overfitting} from within-family hyperparameter selection to controller-class selection.

\paragraph{Gate-and-residual hybrids and hallucination impossibility.} Combining a deterministic gate with a learned fallback recurs across selective generation \citep{chen2023frugalgpt, dekoninck2024unified, jeong2024adaptiverag, asai2024selfrag}. \pith{} singles out the subclass with a \emph{deterministic} rule on a channel disjoint from the fallback's feature block, and gives the regime-theoretic conditions under which this restriction strictly dominates a learned \pit---and when the \pio fallback wins instead. The lattice motivation is sharper where instance-level signal is structurally limited: \citet{kalai2024calibrated} and \citet{karbasi2025impossibility} prove lower bounds on hallucination detection from same-information-source signals; Theorem~\ref{thm:bernstein} provides a finite-sample viability threshold in the same spirit, and Theorem~\ref{thm:partition} identifies when \pio can recover loss in this regime. Mitigation approaches in this literature change decoding \citep{leng2024vcd}, the feature space \citep{farquhar2024semantic}, or the action set \citep{srinivasan2024recoverr, whitehead2022reliable}.

\section{Discussion}
\label{sec:discussion}

Residual mass, sample size, and partition geometry are three genuinely distinct bottlenecks, and which one binds determines which class wins. Each check is computable from a held-out probe and maps to one of the theorems, and the same selection rule places \piz, \pio, \pit, and \pith{} in their predicted regimes across superficially similar benchmarks, arguing that family choice, not within-family refinement, is where most of the loss comes from.

Two boundary conditions sharpen the claims: \pith{} requires a label-free prior, and Theorems~\ref{thm:residual}--\ref{thm:oracle} depend only on $(\alpha, \beta, n, \mathbb{P}(R), \{p_g \gamma_g\})$, so changing the policy pair or judge moves the diagnostics but not the rule. The bottlenecks bind differently across benchmarks (Theorem~\ref{thm:partition} tight on HallusionBench, uncertified \pit on FOLIO, $0.0233$-loss \pith{} gap on TextVQA-OCR); reliable deferred deployment is about matching class to data.

\bibliographystyle{plainnat}
\bibliography{references}

\appendix

\section{Limitations}
\label{app:limitations}

The regime diagnostics are conditional on the chosen action set, loss matrix, feature block, and judge; changing these ingredients can move a benchmark to a different regime, although the same diagnostics can be recomputed. The Bernstein analysis is derived for a selective subproblem and used as a class-level viability proxy for richer multi-action controllers. When the asymptotic ceilings \(C_{\Pi_1}\) and \(C_{\Pi_2}\) are commensurate, the empirical winner can depend on how well the deployed candidate families realize their ceilings; Corollary~\ref{cor:nested}~(iii)'s class prediction is population-optimal but mediated by family capacity at finite~\(n\) (Appendix~\ref{app:synth-pi12}). Semantic-risk labels on HallusionBench and A-OKVQA come from a single judge (InternVL2.5-8B) from a different VLM family than the upstream policies; alternative judges or ensembles could shift the loss matrix and therefore the regime tags. The \pith{} rung further requires a label-free prediction-time prior channel, and our empirical coverage is limited to the benchmarks studied here.

\section{Proofs}
\label{app:proofs}

\subsection{Theorem~\ref{thm:residual} (residual bound)}

Let \(g(x):=\ell(x,a^\star)-\ell(x,\pi(x))\). On \(R^c\), by the definition of the residual set, \(\ell(x,a^\star)\le\ell(x,a)\) for every \(a\in\mathcal{A}\), so \(g(x)\le 0\) on \(R^c\). Therefore
\[
\mathbb{E}[g(x)]
=
\mathbb{E}[g(x)\mathbf{1}_R]+\mathbb{E}[g(x)\mathbf{1}_{R^c}]
\le
\mathbb{E}[g(x)\mathbf{1}_R]
\le
\mathbb{P}(R)\,\sup_{x\in R,a\in\mathcal{A}}\bigl(\ell(x,a^\star)-\ell(x,a)\bigr),
\]
which is the stated bound. \qed

\subsection{Lemma~\ref{lem:necessary} (asymptotic signal condition)}

The Chow-style derivation in \citep{chow2003optimum} and its selective-risk extensions \citep{franc2023optimal, pugnana2023auc} give that the population loss difference of the rejector at coverage \(q\) induced by \(f\), relative to the fixed direct policy, is
\[
\Delta L(q)
=
-q\,(L_w-L_r)\,(\mu_w(q)-\alpha_{\min}),
\]
where \(\mu_w(q)\) is the population precision of the rejected set. Thus the exact improvement condition is \(\mu_w(q)>\alpha_{\min}\). The AUC margin \(\beta=\alpha_{\mathrm{emp}}-\alpha_{\min}\) used in the main text is a scalar proxy for this working-point margin, not an identity between AUC and tail precision. Under the local tail-margin convention stated before Theorem~\ref{thm:bernstein}, the certified loss-reduction scale is \(q^\star\beta(L_w-L_r)\) after optimizing over \(q\in(0,1]\). \qed

\subsection{Theorem~\ref{thm:bernstein} (Bernstein-tight viability threshold)}
\label{app:bernstein}

Let \(m:=\lfloor nq\rfloor\) and let \(\widehat\mu_w(q)\) be the empirical precision on the \(m\) samples with the lowest \(f\)-scores. Conditional on \(f\) (trained on a held-out split under our nested-CV protocol), the bottom-\(q\) labels are independent Bernoulli draws with average success probability \(\mu_w(q)\); per-sample probabilities can vary across the bin, but the empirical-mean variance bound \(\sigma^{2}\le\mu_w(q)(1-\mu_w(q))\) holds for the sample mean either way, and is in turn bounded by \(\alpha(1-\alpha)\) for positive-margin rankers with \(\alpha\ge 1/2\). The deviation bound below thus controls the precision-proxy loss difference \(\widehat\Delta(q)=-q(L_w-L_r)(\widehat\mu_w(q)-\alpha_{\min})\) with \(L_w,L_r\) treated as published constants; the additional fold-wise variance from estimating \(L_w,L_r\) themselves is controlled separately by the action-conditional sample means and is dominated by the Bernoulli term whenever~\eqref{eq:bern} binds.

Bernstein's inequality \citep{boucheron2013concentration, vershynin2018high} in deviation form gives, with probability at least \(1-\delta\),
\[
\bigl|\widehat\mu_w(q)-\mu_w(q)\bigr|
\;\le\;
\sqrt{\tfrac{2\sigma^{2}\log(2/\delta)}{m}}\;+\;\tfrac{2}{3}\cdot\tfrac{\log(2/\delta)}{m}.
\]
The empirical loss difference is \(\widehat\Delta(q)=-q(L_w-L_r)(\widehat\mu_w(q)-\alpha_{\min})\), so \(|\widehat\Delta(q)-\Delta(q)|=q(L_w-L_r)\,|\widehat\mu_w(q)-\mu_w(q)|\). Substituting the Bernstein deviation and using \(m=nq\) together with \(\sigma^{2}\le\alpha(1-\alpha)\), the condition \(|\Delta(q)|\ge|\widehat\Delta(q)-\Delta(q)|\) (which suffices for \(\widehat\Delta(q)<0\) with probability at least \(1-\delta\)) becomes, after cancelling \(q(L_w-L_r)\),
\[
\beta
\;\ge\;
\sqrt{\tfrac{2\alpha(1-\alpha)\log(2/\delta)}{nq}}\;+\;\tfrac{2}{3}\cdot\tfrac{\log(2/\delta)}{nq}.
\]
The dominant term on the right is of order \(1/\sqrt{n}\) and the sub-leading term is of order \(1/n\). Dropping the sub-leading term and squaring the resulting inequality \(\beta^{2}\ge 2\alpha(1-\alpha)\log(2/\delta)/(nq)\) gives the leading-order sufficient condition \(n\ge 2\alpha(1-\alpha)\log(2/\delta)/(q\beta^{2})\), which is precisely~\eqref{eq:bern}. The dropped correction is of lower order in \(n\) and does not change the qualitative regime classification (Section~\ref{sec:experiments} reports per-benchmark instantiations). \qed

\subsection{Proposition~\ref{prop:lower} (information-theoretic lower bound)}
\label{app:lower}

The sign of the loss difference \(\Delta(q)\) is determined by whether the bottom-\(q\) precision \(\mu_w(q)\) exceeds \(\alpha_{\min}\). Any test that determines this sign from \(n\) i.i.d.\ samples must in particular solve the binary hypothesis test \(H_0\!:\mu_w=\alpha_{\min}\) versus \(H_1\!:\mu_w=\alpha_{\min}+\beta\) using the \(m=\lfloor nq\rfloor\) Bernoulli observations (wrong/right labels) in the rejected set.

Let \(P_0=\mathrm{Bern}(p)\), \(P_1=\mathrm{Bern}(p+\beta)\), with \(p=\alpha_{\min}\). For any test \(\varphi\), the Bretagnolle--Huber form of Le Cam's inequality gives
\[
P_0^{\otimes m}(\varphi=1)+P_1^{\otimes m}(\varphi=0)
\;\ge\;
\tfrac12\exp\!\bigl(-m\,\mathrm{KL}(P_0\|P_1)\bigr).
\]
If both type-I and type-II errors are at most \(\delta\), the left-hand side is at most \(2\delta\), hence
\[
m\,\mathrm{KL}(P_0\|P_1)
\;\ge\;
\log\!\frac{1}{4\delta}.
\]
A standard local Bernoulli KL bound gives, for \(0<\beta\le \tfrac12\min\{p,1-p\}\),
\[
\mathrm{KL}\bigl(\mathrm{Bern}(p)\,\|\,\mathrm{Bern}(p+\beta)\bigr)
\;\le\;
\frac{2\beta^{2}}{p(1-p)}.
\]
Combining the two displays gives
\[
m
\;\ge\;
\frac{p(1-p)}{2\beta^{2}}\log\!\frac{1}{4\delta}.
\]
It remains to relate \(p(1-p)\) to \(\alpha(1-\alpha)\). Under the proposition's assumption \(\beta\le \tfrac12\min\{p,1-p\}\), we have \(\alpha = p+\beta \le 3p/2\) and \(1-\alpha = 1-p-\beta \ge (1-p)/2\); together with \(\alpha\ge p\) and \(1-\alpha\le 1-p\), this gives \(\alpha(1-\alpha)\in[p(1-p)/2,\,(3/2)p(1-p)]\). Hence \(p(1-p)\) and \(\alpha(1-\alpha)\) differ by at most a factor of three, and substituting \(m=\lfloor nq\rfloor\) yields~\eqref{eq:lower} after adjusting the universal constant \(c\). \qed

\subsection{Theorem~\ref{thm:partition} (distribution-dependent partition lower bound)}

For the population lower bound, condition on the cell \(G_g\) and observe that on each cell the partition router plays \(a_g^\star\), the cell-optimal action, so its conditional loss is \(\mathbb{E}[\ell(x,a_g^\star)\mid x\in G_g]\). Averaging over cells,
\[
\Dpart(\mathcal{G})
=\sum_g p_g\bigl(\mathbb{E}[\ell(x,a^\star)\mid x\in G_g]-\mathbb{E}[\ell(x,a_g^\star)\mid x\in G_g]\bigr)
=\sum_g p_g\gamma_g.
\]
Each summand is non-negative because \(a_g^\star\) minimizes conditional loss on \(G_g\), so \(\Dpart(\mathcal{G})\ge\max_g p_g\gamma_g\), which is~\eqref{eq:partition}.

For the sample-complexity statement, condition on the partition and let \(M_g\sim\mathrm{Binom}(n_{\mathrm{tr}},p_g)\) be the realized training-fold count on cell \(G_g\), with \(n_{\mathrm{tr}}=(\kappa-1)n/\kappa\) and mean \(\bar m_g=n_{\mathrm{tr}}p_g\). Because Bernstein's bound is convex in \(1/m_g\), substituting \(\bar m_g\) directly is not rigorous; we first lower-bound \(M_g\) by a multiplicative Chernoff inequality, \(\mathbb{P}(M_g\ge \bar m_g/2)\ge 1-\exp(-\bar m_g/8)\), then condition on this event with \(\delta/2\) budget. On it, Bernstein's inequality applied to each action's empirical mean on cell \(G_g\) with variance at most \(\sigma^{2}\) and range at most \(L_{\max}\) gives
\[
\mathbb{P}\bigl(|\widehat L_g(a)-L_g(a)|\ge t\bigr)
\le
2\exp\!\left(-\frac{(\bar m_g/2)\,t^{2}}{2\sigma^{2}+(2/3)L_{\max}t}\right).
\]
Union-bounding over the \(|\mathcal{A}|\) actions with the remaining \(\delta/2\) budget, the empirical argmin's Bernstein deviation is at most \(\gamma_g/2\) on every action whenever
\(
\bar m_g
\ge
(16\sigma^{2}+(8/3)L_{\max}\gamma_g)\log(4|\mathcal{A}|/\delta)/\gamma_g^{2},
\)
in which case \(\widehat L_g(\widehat a_g^\star)\le \widehat L_g(a_g^\star)\le L_g(a_g^\star)+\gamma_g/2\) and therefore \(L_g(\widehat a_g^\star)\le L_g(a_g^\star)+\gamma_g=L_g(a^\star)\), so the empirical \pio estimator's cell loss is no worse than the global baseline on cell \(G_g\). Substituting \(\bar m_g=(\kappa-1)np_g/\kappa\) and rearranging gives the stated bound. Strictly, the requirement is
\(
\bar m_g\ge\max\!\left\{8\log(2/\delta),\;(16\sigma^{2}+(8/3)L_{\max}\gamma_g)\log(4|\mathcal{A}|/\delta)/\gamma_g^{2}\right\},
\)
where the first term is the Chernoff floor on cell occupancy and the second the Bernstein concentration term, matching the explicit max in the theorem statement; the Bernstein term dominates whenever \(\gamma_g\) is bounded away from \(L_{\max}\). \qed

\subsection{Theorem~\ref{thm:oracle} (held-out selection bound)}
\label{app:oracle}

Let \(M=|\mathcal{M}|\) be the number of candidate classes. For each class \(m\in\mathcal{M}\), the inner \(\kappa\)-fold CV estimate \(\widehat R_{\mathrm{CV}}(m)\) is the average of held-out-fold empirical risks, with each held-out fold containing \(n_{\mathrm{in}}=n_{\mathrm{out}}/\kappa\) samples and loss bounded in \([0,L_{\max}]\). Treating the fold-trained estimator as fixed conditional on its training split, Hoeffding's inequality gives the displayed deviation rate for each held-out fold; averaging over folds preserves the same order. Thus, for the inner-CV estimand \(\bar R_\kappa(m)\),
\[
\mathbb{P}\bigl(|\widehat R_{\mathrm{CV}}(m)-\bar R_\kappa(m)|>t\bigr)
\;\le\;
2\exp\!\Bigl(-\frac{2\,n_{\mathrm{in}}\,t^{2}}{L_{\max}^{2}}\Bigr),
\]
where \(\bar R_\kappa(m):=\mathbb{E}_{S'}[R(\widehat\pi_m(S'))]\) is the expected risk of the estimator trained on a single inner-training fold of size \(n_{\mathrm{out}}(\kappa-1)/\kappa\). A union bound over the \(M\) candidate classes gives
\[
\mathbb{P}\Bigl(\max_{m}|\widehat R_{\mathrm{CV}}(m)-\bar R_\kappa(m)|>t\Bigr)
\;\le\;
2M\exp\!\Bigl(-\frac{2\,n_{\mathrm{in}}\,t^{2}}{L_{\max}^{2}}\Bigr).
\]
Define the threshold \(t^\star:=L_{\max}\sqrt{\log(2M)/(2n_{\mathrm{in}})}\), at which the right-hand side of the union bound equals \(1\). On the event that all class deviations are at most some \(t\), the selection \(\widehat m=\arg\min_m\widehat R_{\mathrm{CV}}(m)\) satisfies
\[
\bar R_\kappa(\widehat m)
\;\le\;
\widehat R_{\mathrm{CV}}(\widehat m)+t
\;\le\;
\widehat R_{\mathrm{CV}}(m^\star)+t
\;\le\;
\bar R_\kappa(m^\star)+2t
\]
for any competitor class \(m^\star\). Taking expectations through tail integration \(\mathbb{E}[Z_+]=\int_0^{\infty}\mathbb{P}(Z>x)\,\mathrm{d}x\) applied to \(Z=\bar R_\kappa(\widehat m)-\min_m \bar R_\kappa(m)\): the integrand is at most \(1\) for \(t\le t^\star\) (contributing \(2t^\star\) after the factor-of-\(2\) chaining above) and at most \(2M\exp(-2n_{\mathrm{in}}t^{2}/L_{\max}^{2})\) for \(t>t^\star\) (a Gaussian tail bounded by \(L_{\max}/(2\sqrt{2n_{\mathrm{in}}\log(2M)})\le t^\star\) for \(M\ge 2\)). Both contributions are at most \(2t^\star\), giving
\begin{align*}
\mathbb{E}[\bar R_\kappa(\widehat m)]
\;\le\;
&\min_{m\in\mathcal{M}}\bar R_\kappa(m)+4t^\star
\;=\;
\min_{m\in\mathcal{M}}\bar R_\kappa(m)+2\sqrt{2}\,L_{\max}\sqrt{\frac{\log(2|\mathcal{M}|)}{n_{\mathrm{in}}}}\\
\;\le\;
&\min_{m\in\mathcal{M}}\bar R_\kappa(m)+4L_{\max}\sqrt{\frac{\log(2|\mathcal{M}|)}{n_{\mathrm{in}}}}.
\end{align*}
If refitting a selected class from inner-training size to the full outer-training size changes its expected risk by at most \(\varepsilon_{\mathrm{stab}}(n_{\mathrm{out}},\kappa)\), and the same holds for the best competitor class, the refit bound follows by adding \(2\varepsilon_{\mathrm{stab}}(n_{\mathrm{out}},\kappa)\). \qed

\section{Setup and implementation}
\label{app:setup}

This appendix details the implementation. Section~\ref{app:features} specifies the per-benchmark feature blocks; Section~\ref{app:loss} the loss matrix and the diagnostic constants $L_r,L_w,L_a$; Section~\ref{app:l2d} the controller family pool and the cost-sensitive L2D adaptations; Section~\ref{app:protocol} the strict nested-CV protocol and the retrieve-action prompting.

\subsection{Feature block specification}
\label{app:features}

For HallusionBench and A-OKVQA the controller consumes a fixed scalar feature block of 41 dims (HallusionBench) or 39 dims (A-OKVQA), built entirely from the upstream Qwen2.5-VL prediction-time outputs (per-action margins, agreements, sampled-completion statistics, modality-bypass probes, and self-verification probes); the InternVL2.5-8B judge is used \emph{only} to populate the semantic-risk channel of the loss matrix (Appendix~\ref{app:loss}) and never enters the feature block. FOLIO uses a 12-dimensional simplified block constructed from the same Qwen2.5-VL outputs in text-only mode (described at the end of this subsection). SMS-Spam uses a 12-dimensional TF-IDF--based block (Appendix~\ref{app:sms-spam}). The TextVQA-OCR add-on uses a separate non-OCR feature block (exposed to \(\piz,\pio,\pit\); OCR-derived features go only to \pith): three scalars (direct-confidence, question length in tokens, number of image classes) plus one-hot indicators (the first question token, the first 12 question tokens, tokens matching a small fixed text-key vocabulary, and the first 8 image-class names), assembled via a sparse \texttt{DictVectorizer} so the realized width is vocabulary-dependent. The \pith{} prior block additionally exposes OCR-derived signals (OCR confidence/margin, prediction length and digit/year flags, question/prediction overlap, training-frequency log) under the deterministic OCR-confidence gate. The hallucination/VQA block is organized into six groups, each of which expands to multiple scalars:

\begin{enumerate}[leftmargin=1.6em]
\item \emph{Per-action confidence and margin} (7 dims). For each of the three non-abstain actions (direct, retrieve, defer): the greedy softmax margin and the mean per-token log-probability of the generated answer. Additionally, a context-aware margin for the retrieval action. These features capture first-order output-distribution signals for every candidate action.
\item \emph{Cross-action agreement} (3 dims). Pairwise token-level agreement between the answers produced by direct, retrieve, and defer (\texttt{agree\_dr}, \texttt{agree\_de}, \texttt{agree\_re}; the suffix letters are implementation identifiers and predate this paper's terminology).
\item \emph{Question metadata} (2 dims). Question length in words and cosine similarity between question and context embeddings.
\item \emph{Semantic similarity and benchmark categories} (8 dims on HallusionBench, 10 on A-OKVQA). Pairwise cosine similarities among the sentence embeddings of the three action-specific answers and between each answer and the question (6 pairs), plus benchmark-specific categorical one-hot indicators (HallusionBench: 2 dims, VD/VS category; A-OKVQA: 4 dims, question-type).
\item \emph{Stochastic-sampling uncertainty} (8 dims, from \(K{=}10\) sampled completions). Self-consistency count ratio (max-answer frequency divided by \(K\)), sample disagreement, Shannon semantic entropy over parsed distinct answers, predictive entropy from the first-token softmax, and means and standard deviations of the per-sample softmax margin and sequence log-probability across the \(K\) draws.
\item \emph{Verification and modality-bypass probes} (13 dims on HallusionBench, 9 on A-OKVQA). Five scalars from a modality-bypass (image-blanked) probe applied to the vision benchmarks: a binary disagreement flag, softmax margin, sequence log-probability, \(P(\text{yes}){-}P(\text{no})\), and top-1 token probability. Plus four scalars from a 3B self-verification re-ask (\(P(\text{yes})\), \(P(\text{no})\), their difference, and a binary text-contains-yes flag); on HallusionBench this is repeated under the 7B re-asker, contributing four further scalars (\(P(\text{yes})\), \(P(\text{no})\), their difference, and the binary text-contains-yes flag).
\end{enumerate}

All features are computed once per sample on the upstream pipeline and represented as a dense matrix paired with the loss matrix. Features are standardized per outer-training fold via \texttt{sklearn.preprocessing.StandardScaler}, and all AUCs reported in Table~\ref{tab:theory} are computed with \texttt{sklearn.metrics.roc\_auc\_score} on the concatenation of outer-test folds so that the reported \(\alpha_{\mathrm{emp}}\) reflects the same data budget the controllers see.

The FOLIO 12-d block is built from groups~1--2 above (no $K{=}10$ stochastic, no blind-probe, no self-verification, no InternVL features, no question metadata) plus two predicted-label indicators (\texttt{pred\_true}, \texttt{pred\_uncertain}): per-action confidence margin and sequence log-prob (6 dims), three pairwise agreements (3 dims), the recover-context margin (1 dim), and the two indicator features (2 dims). This deliberately constrained block is part of the variance-bounded regime instantiation: Theorem~\ref{thm:bernstein}'s threshold $n_{\min}\propto \alpha(1{-}\alpha)/\beta^2$ depends on the AUC the feature block can extract, so a thinner block at small $n$ instantiates the Cor~\ref{cor:nested}~(ii) sub-case where \pit's sign is uncertified.

\subsection{Loss matrix construction}
\label{app:loss}

\paragraph{Loss components and per-benchmark labeling.} The per-sample loss \(\ell(x_i,a)\) is constructed from three operational quantities. Correctness \(c(x_i,a)\in\{0,1\}\) is evaluated against the benchmark gold label. Operational cost \(k(x_i,a)\) is measured in action-cost units (1.0 for direct, 2.0 for retrieve, 2.3 for defer, 0.0 for abstain). Semantic risk \(h(x_i,a)\in\{0,0.5,1\}\) is a per-action grounded-support label: \(h{=}0\) (grounded), \(0.5\) (unverifiable), \(1.0\) (contradicted), with \(h{=}0\) for $a_{\mathrm{abstain}}$ by construction. Labeling is benchmark-specific and run offline. \emph{HallusionBench} and \emph{A-OKVQA}: each (sample, action) candidate answer's most specific visual proposition is extracted and classified against image evidence by an InternVL2.5-8B judge as grounded / unverifiable / contradicted. \emph{FOLIO} uses a rule-based proxy ($h{=}0.5$ if the action's answer disagrees with the gold label, $h{=}0$ if it agrees), keeping the loss matrix simple while still exposing wrong answers to a calibrated penalty; using a learned NLI judge here would conflate the controller's behaviour with the judge's quality on logical inference. \emph{SMS-Spam} uses canonical-cost weighting without a separate judge (Appendix~\ref{app:sms-spam}).

\paragraph{Headline formula and weight robustness.} The headline core loss is
\[
\ell(x_i,a)=1.0\cdot(1-c(x_i,a))+1.0\cdot h(x_i,a)+0.05\cdot k(x_i,a),
\]
which is the exact formula used for the four core benchmarks throughout this paper. Abstention therefore has loss \(1.0\), because it is treated as an always-incorrect zero-risk zero-cost action. Since this weight vector is a cost-sensitive operating point, Table~\ref{tab:weight-sensitivity} reruns the deployable \(\piz/\pio/\pit\) pool under local one-factor perturbations of these weights: \(w_{\mathrm{c}}\in\{0.75,1.0,1.25\}\), \(w_{\mathrm{h}}\in\{0.75,1.0,1.25\}\), and \(w_{\mathrm{k}}\in\{0.025,0.05,0.10\}\), varying one coordinate at a time around \((1,1,0.05)\). The TextVQA-OCR add-on instead uses answer loss \(1-\)soft-VQA accuracy because the experiment isolates OCR copying rather than semantic-risk/cost tradeoffs.

\paragraph{Diagnostic constants and numerical examples.} The quantities \(L_r:=\mathbb{E}[\ell(x,a_{\mathrm{direct}})\mid c_{\mathrm{direct}}{=}1]\), \(L_w:=\mathbb{E}[\ell(x,a_{\mathrm{direct}})\mid c_{\mathrm{direct}}{=}0]\), and \(L_a:=\mathbb{E}[\ell(x,a_{\mathrm{abstain}})]\) are computed as simple sample means over the respective subsets of the full core benchmark for the \pit viability diagnostic; they are global (not fold-specific) constants for each benchmark, so \(\alpha_{\min}=(L_a-L_r)/(L_w-L_r)\) is a single number per benchmark that does not depend on the CV fold. For scale, the \texttt{always\_direct} losses decompose under this formula as HallusionBench \(1.0848=(1-0.6685)+0.7033+0.0500\), A-OKVQA \(0.4138=(1-0.8306)+0.1944+0.0500\), and FOLIO \(0.7520=(1-0.5320)+0.2340+0.0500\). Correctness \(c\) is gold-evaluated and therefore identical across judge choices; on HallusionBench and A-OKVQA the per-action risk \(h\) values entering Table~\ref{tab:main} come from the InternVL2.5-8B judge from a different VLM family than the upstream Qwen2.5-VL policies, so the judge does not score outputs from its own family. On HallusionBench the population-mean per-action losses are $(1.085, 1.136, 1.104, 1.000)$ for (direct, retrieve, defer, abstain), so abstain becomes the best fixed action by a wide margin and the \texttt{fair\_fixed\_train} CV variant reliably recovers it; on A-OKVQA and FOLIO direct remains the best fixed action on outer-test. Table~\ref{tab:main}'s $\piz$ column reports the strict-CV-selected variant on each benchmark.

\subsection{Family pool and baseline implementations}
\label{app:l2d}\label{app:tree}

Our family pool is evaluated through a uniform \texttt{fit(X, L, C)}/\texttt{predict(X)} interface in the strict nested CV pipeline. \piz holds \texttt{always\_direct} and a fair-fixed CV variant. \pio (KMeans) uses KMeans routers with \(K\in\{4,5,6,8\}\) on the standardized feature block, with per-cell action assigned by conditional loss argmin on outer-train and a global-best fallback for cells with fewer than three training samples. An alternative \pio (CART) realization uses \texttt{DecisionTreeClassifier} with \texttt{max\_depth}\(\in\{3,4\}\) and \texttt{min\_samples\_leaf}\(=5\), fit on per-sample loss-argmin labels with the same per-leaf override and small-cell fallback (Table~\ref{tab:tree}). \pit uses \texttt{HistGradientBoostingClassifier} (\texttt{max\_depth}\(\in\{3,4\}\)) and a calibrated logistic plug-in (\(C{=}0.3\), referred to as \texttt{Selective}-\(C{=}0.3\) elsewhere in this paper), with each family predicting a four-action argmin from the feature block.

We additionally evaluate two cost-sensitive adaptations of canonical learning-to-defer baselines through the same interface. The Mozannar--Sontag adaptation \citep{mozannar2020consistent} is a row-replicated multinomial logistic regression: each training row \(x_i\) is replicated \(|\mathcal{A}|=4\) times with the action index as the label and a sample weight \(w_{i,a}=L_{\max,i}-L[i,a]\) encoding the cost margin, with \(L_2\) regularization \(C\in\{0.3, 1.0\}\); at test time we predict the softmax argmax. The Narasimhan adaptation \citep{narasimhan2022posthoc} is a post-hoc plug-in: one \texttt{HistGradientBoostingRegressor} per action (\texttt{max\_iter}\(=200\), \texttt{max\_depth}\(\in\{3,4\}\), \texttt{learning\_rate}\(=0.05\)) is trained on the per-action loss column, and the test policy is \(\pi(x)=\arg\min_a\widehat L_a(x)\). Both are loss-matrix-compatible cost-sensitive instantiations rather than reproductions of the original human-deferral objectives; their detailed strict-CV results are in Appendix~\ref{app:l2d-results}.

\subsection{Evaluation protocol and reproducibility}
\label{app:protocol}

\paragraph{Strict nested CV.} All headline tables use strict nested 5-fold-by-5-seed cross-validation with family selection on outer-train only, followed by a refit of the selected family on outer-train only and a single evaluation on outer-test. Non-strict variants that refit on outer-train-plus-outer-test are excluded from the canonical results: in pre-paper diagnostic runs on HallusionBench's best \pit family the non-strict refit produced losses inflated by ${\sim}0.1$ relative to the strict variant, the textbook overfit signature of refitting on the evaluation fold. Families, hyperparameter grids, and seeds are fixed; the strict outer CV is deterministic given the fixed seeds and pre-computed loss matrix.

\paragraph{Retrieve action.} The \(a_{\mathrm{retrieve}}\) action on HallusionBench, A-OKVQA, and FOLIO is generated by the same upstream model (Qwen2.5-VL-3B-Instruct) under a grounded-context prompting pass: a single forward call instructs the model first to enumerate scene context literally (objects, attributes, relations on visual benchmarks; key facts and entities on FOLIO) and then to commit to the question's answer conditional on that enumeration, with the final commitment extracted by deterministic regex from the response. There is no external retrieval corpus; the ``retrieved'' context is the model's own free-form description, so \(a_{\mathrm{retrieve}}\) costs one extra forward pass over \(a_{\mathrm{direct}}\) but draws on no information unavailable to the upstream model.

\paragraph{Hardware.} All experiments run on AMD EPYC 7543 CPU cores and a single NVIDIA A100 GPU.

\section{Per-benchmark details}
\label{app:per-bench}

Each subsection below applies Corollary~\ref{cor:nested} to one core benchmark, starting with the predicted regime and then giving empirical detail beyond the main text.

\subsection{SMS-Spam: residual-bounded witness for Theorem~\ref{thm:residual}}
\label{app:sms-spam}

The SMS-Spam Collection~\citep{almeida2011smsspam} is a 5{,}574-message English binary classification corpus (ham vs.\ spam). We use a deterministic 80/20 split (\(n_{\mathrm{train}}{=}4460\), \(n_{\mathrm{test}}{=}1114\); seed-0 NumPy permutation) and CPU-only TF-IDF features. The three deployable actions are: \emph{direct} = TF-IDF (1,2)-gram (max\_features \(50{,}000\), min\_df \(2\)) + calibrated LinearSVC (\(C{=}1.0\), 3-fold sigmoid calibration), \(0.9910\) test accuracy; \emph{retrieve} = TF-IDF (1,1)-gram (max\_features \(20{,}000\), min\_df \(2\)) + KNN (\(k{=}10\), cosine), \(0.9811\); \emph{defer} = same TF-IDF (1,1)-gram + logistic regression (\(C{=}1.0\)), \(0.9758\). The 12-d scalar feature block is per-action top-1 minus top-2 probability, per-action top-1 log-probability, three pairwise prediction agreements, and 3 TruncatedSVD components fit on the SMS training split's (1,2)-gram TF-IDF matrix and applied to the held-out split. Loss uses canonical core weights \((1,1,0.05)\) and the same \(\{1.0, 2.0, 2.3, 0.0\}\) action-cost vector as the other core benchmarks; abstain has zero operational cost \(k\) and (per the convention in Section~\ref{sec:setup}) loss \(1.0\) under the canonical correctness weight.

This configuration deliberately instantiates the residual-bounded regime: the bigram + calibrated SVC direct is task-specialized, while the unigram logistic-regression defer is generic and not stronger on this task. We measure the four diagnostics from Section~\ref{sec:theory} and Theorem~\ref{thm:residual}: \(\mathbb{P}(R){=}0.009\), oracle \(\Pi_2\) gain \({=}0.003\), oracle \(\Pi_1\) partition gain \({=}0.000\) (over a post-hoc \(K\in\{2,3,4,5,6,8\}\) KMeans sweep on an SVD-8 diagnostic embedding), and the local tail-margin condition of Lemma~\ref{lem:necessary} fails: \(\mathbb{P}(R) = 0.009\) caps the bottom-\(q\) precision by \(\mu_w(q) \le \mathbb{P}(R)/q \le 0.03\) for \(q \ge 0.3\), below \(\alpha_{\min} = 0.95\) (abstain fallback) and \(\alpha_{\min} = 0.089\) (defer fallback), so no rejector improves on \(a_{\mathrm{direct}}\) asymptotically. We record this as \(\beta \le 0\) in Table~\ref{tab:theory} under the AUC scalar-proxy convention. All four are below the residual-bounded thresholds \((<0.02, <0.005, <0.005)\). Theorem~\ref{thm:partition} sharpens the \pio diagnostic: \(\max_g p_g\gamma_g{=}0\) means every cluster's cell-best action coincides with the global-best \(a_{\mathrm{direct}}\), so \(\pi_1^\star \in \piz\) at the population level. Under strict nested 5-fold-by-5-seed CV, every \pio family (KMeans-\(K\in\{4,5,6,8\}\)) collapses to direct (\(\Delta = +0.0000\)), the best \pit family is \texttt{SelectiveCalibrated\_C0.3} at \(0.0591\pm0.0000\) (\(\Delta = +0.0002\)), and the two HGBC families lose by \(\Delta\in\{+0.0007, +0.0011\}\) to the fixed action---the predicted ordering \(\piz \le \pio \le \pit\) up to multi-seed noise.

A simple structural lower bound on \(\mathbb{P}(R)\) explains why this regime is rare under standard deferral setups. If defer's accuracy exceeds direct's by \(\delta>0\), then \(\mathbb{P}(R)\ge\delta\) (the direct-wrong/defer-correct samples already form a \(\delta\)-fraction). Under the canonical defer cost premium \(0.05(c_{\mathrm{defer}}-c_{\mathrm{direct}})=0.065\), residual loss is bounded only when \(\delta\) is comparable to or below this premium. The other three core benchmarks all have substantial residual mass: HallusionBench (defer-direct accuracy gap \(\delta{=}0.037\)) and A-OKVQA (\(\delta{=}0.034\)) violate the structural lower bound by defer's accuracy advantage; FOLIO (\(\delta{=}-0.015\), defer slightly weaker) instead has \(\mathbb{P}(R){=}0.175\) driven by partition-level abstain dominance on its Uncertain subgroup. All three sit outside Theorem~\ref{thm:residual}'s scope; SMS-Spam, with a generic and weaker fallback that produces neither a defer-side accuracy advantage nor a partition-level abstain region, satisfies it. The witness is therefore not a property of the corpus alone but of the cost-tiered configuration in which the fallback is non-specialized---which is the deployment regime Theorem~\ref{thm:residual} is meant to characterize.

\subsection{HallusionBench: instance-level wins on the high-signal Bernstein-viable regime}
\label{app:cluster}

\begin{figure}[t]
  \centering
  \includegraphics[width=0.55\linewidth]{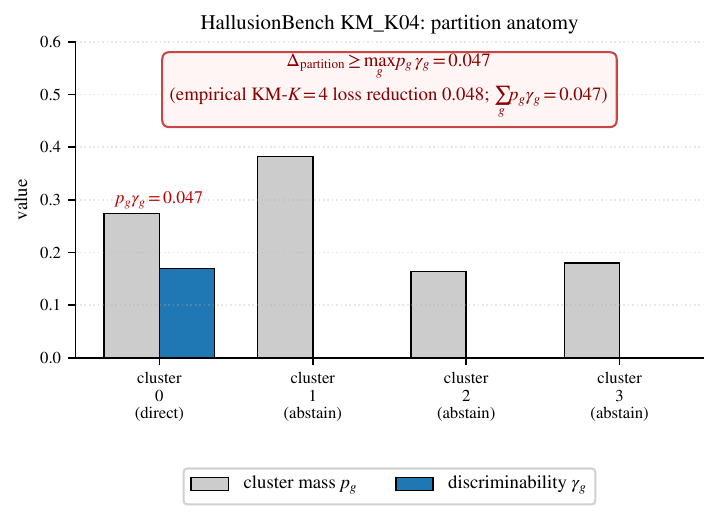}
  \caption{Cluster anatomy of HallusionBench KM-\(K{=}4\), seed \(42\), witnessing Theorem~\ref{thm:partition}. Under the canonical loss matrix the global best fixed action is abstain; only cluster~0 (\(p_0{=}0.274\)) carries positive discriminability by preferring direct (\(\gamma_0{=}0.170\)), contributing \(p_0\gamma_0{=}0.047\). The empirical strict-CV KMeans-\(K{=}4\) loss reduction is \(0.048\), within sampling noise of \(\sum_g p_g\gamma_g{=}0.047\); the bound is empirically tight at \(K{=}4\) on HallusionBench.}
  \label{fig:cluster}
\end{figure}

Under the InternVL2.5-8B judge, HallusionBench at the full \(n{=}920\) corpus is comfortably above the Bernstein threshold (\(n_{\min}^{\mathrm{Bern}}{=}23\), with \(n\beta^2{=}202\); Theorem~\ref{thm:bernstein}); both \pio (best KMeans-\(K{=}8\), \(0.091\) loss reduction) and \pit (best HGBC-\texttt{md3}, \(0.103\) loss reduction) beat the fixed action, with \pit edging \pio by \(\approx 3\) seed-sd of \pit. The deployable winner is therefore \pit; \pio remains a useful lower-complexity diagnostic for Theorem~\ref{thm:partition}. Figure~\ref{fig:cluster} reports the partition anatomy at the simplest router (\(K{=}4\), seed \(42\)): InternVL2.5-8B's stricter judging of direct-wrong samples makes \(a_{\mathrm{abstain}}\) the global best fixed action under the canonical loss matrix (mean direct loss \(1.085\) vs.\ abstain's \(1.000\)), so by the definition \(\gamma_g = \mathbb{E}[\ell(x, a^\star) - \ell(x, a_g^\star) \mid x \in G_g]\) a cluster contributes positively only when its cell-best action differs from \(a^\star{=}a_{\mathrm{abstain}}\). One cluster carries essentially all the partition gain at \(K{=}4\): cluster~0 (\(p_0{=}0.274\)) prefers \(a_{\mathrm{direct}}\) with \(\gamma_0{=}0.170\), contributing \(p_0\gamma_0{=}0.047\); the remaining three clusters all prefer \(a_{\mathrm{abstain}}\) and contribute zero, so \(\max_g p_g\gamma_g=\sum_g p_g\gamma_g=0.047\). The strict-CV empirical KM-\(K{=}4\) loss reduction is \(0.048\pm0.005\) relative to the \texttt{fair\_fixed\_train} baseline of \(1.000\), within sampling noise of \(\sum_g p_g\gamma_g\): Theorem~\ref{thm:partition}'s lower bound is empirically tight at \(K{=}4\). The instance-level recoverable margin (HGBC-\texttt{md3} at \(0.103\)) exceeds this partition ceiling, so \pit wins on HallusionBench by Corollary~\ref{cor:nested} item~(iii)'s ceiling comparison, with the partition diagnostic confirming that \pio's gain is concentrated in a single direct-preferring subgroup rather than diffused across the partition.

\subsection{A-OKVQA: deployable \texorpdfstring{\pit}{Pi\_2} winner with \texorpdfstring{\pio}{Pi\_1} boundary}
\label{app:aokvqa}

A-OKVQA at full \(n{=}1145\) is the deployable \(\pit\) winner (Table~\ref{tab:main}, Selective $C{=}0.3$ at \(-0.033\) versus best \pio KMeans-$K{=}4$ at \(-0.024\)), but with the high-signal \pio/\pit auto-pick boundary characteristic of Corollary~\ref{cor:nested} item~(iii) when the two ceilings are commensurate: $C_{\Pi_2}=q^\star\beta(L_w-L_r)\approx 0.094$ vs.\ $C_{\Pi_1}=\max_K\sum_g p_g\gamma_g\approx 0.039$ predicts \pit, but the recovered empirical margins are within one seed-sd of each other and inner-CV splits \pio:\pit as $10{:}15$ over $25$ outer-folds (Appendix~\ref{app:auto-select}). To complement this deployable result, Appendix~\ref{app:pi3-rationale} reports a gold-rationale oracle that uses an answer-derived channel and is therefore not deployable; it pins down the \pith-style ceiling that an external prior could in principle reach.

\subsection{A-OKVQA: gold-rationale \texorpdfstring{\pith}{Pi\_3} oracle (non-deployable)}
\label{app:pi3-rationale}

A-OKVQA contains three human-written rationales per sample, outside the v10 scalar block. Because these rationales were collected after the answers and ask workers to explain why the correct answer is correct \citep{schwenk2022aokvqa}, they are answer-derived and not deployable side information. We treat the following as a gold-rationale \emph{oracle experiment}: it measures the upper-bound value of an answer-derived explanatory channel, not a deployable \pith result.

We define a parameter-free binary gate \(\rho(x)\) that fires when any content word of the model's direct answer appears in any rationale of \(x\). On A-OKVQA \(\rho\) fires on \(806/1145{=}70.4\%\) of samples with empirical direct-correct rate \(94.4\%\) inside the fired set (vs.\ base rate \(83.1\%\)); on the remaining \(29.6\%\) the rate collapses to \(56.0\%\), giving AUC \(0.784\). The \pith-style family plays \(a_{\mathrm{direct}}\) when \(\rho{=}1\) and uses a KMeans-\(K{=}4\) fallback on the non-fire subset, with no tunable hyperparameters.

\begin{table}[t]
  \centering
  \small
  \setlength{\tabcolsep}{4pt}
  \renewcommand{\arraystretch}{0.95}
  \begin{tabular}{lccc}
    \toprule
    A-OKVQA (\(n{=}1145\)), best \pit per block & \texttt{loss} & \texttt{std} & \(\Delta\) vs \pith \\
    \midrule
    \multicolumn{4}{l}{\emph{Canonical v10 scalar block (39-d), no rationale access}}\\
    \piz: always\_direct                                  & \(0.4138\) & \(0.0000\) & \(+0.0584\)\\
    \pio: KMeans-\(K{=}4\)                                  & \(0.3902\) & \(0.0085\) & \(+0.0348\) (4.0 seed-sd)\\
    \pit: SelectiveCalibrated \(C{=}0.3\)                 & \(0.3805\) & \(0.0032\) & \(+0.0251\) (6.8 seed-sd)\\
    \midrule
    \multicolumn{4}{l}{\emph{\pit given progressively richer rationale representations}}\\
    \pit HGBC-\texttt{md3} + exact-match (40-d)            & \(0.3749\) & \(0.0052\) & \(+0.0195\) (3.5 seed-sd)\\
    \pit HGBC-\texttt{md3} + text-overlap stats (45-d)     & \(0.3737\) & \(0.0045\) & \(+0.0183\) (3.7 seed-sd)\\
    \pit HGBC-\texttt{md3} + TF-IDF top-200 (239-d)        & \(0.3847\) & \(0.0043\) & \(+0.0293\) (6.2 seed-sd)\\
    \midrule
    \multicolumn{4}{l}{\emph{\pith-style gold-rationale oracle (not a deployable feature)}}\\
    \pith oracle: rationale gate + KM-\(K{=}4\) fallback    & \(\mathbf{0.3554}\) & \(\mathbf{0.0019}\) & \(\mathbf{0\ (oracle)}\)\\
    \bottomrule
  \end{tabular}
  \caption{Gold-rationale oracle decomposition on A-OKVQA (5-seed strict CV). \textbf{Top:} scalar-only deployable families. \textbf{Middle:} \pit HGBC given three progressively richer learned representations of the rationale channel. \textbf{Bottom:} the deterministic gold-rationale gate plus KMeans fallback. None of the three augmented \pit variants closes the gap to the oracle gate (\(3.5\)--\(6.2\) seed-sd above), and each row reports the lowest-mean \pit family within its block.}
  \label{tab:pi3-real}
\end{table}

Under strict 5-fold-by-5-seed CV the oracle reaches \(0.3554\pm0.0019\), beating the best scalar \pio at \(0.3902\pm0.0085\) by \(-0.0348\) and the best scalar \pit at \(0.3805\pm0.0032\) by \(-0.0251\), gaps of about \(4\) and \(7\) seed-sd units, respectively. The result is numerically consistent with the region where \pith wins in Appendix~\ref{app:pi3-synth}; the gate's AUC matches the high-prior-signal region. However, the channel itself is answer-derived, so the experiment serves as motivation for label-free \pith channels (OCR, retrieval, knowledge bases) rather than a deployable A-OKVQA \pith result.

\subsection{FOLIO: variance-bounded witness for Corollary~\ref{cor:nested}~(ii)}
\label{app:folio}

FOLIO~\citep{han2024folio} is a 203-sample first-order-logic NLI benchmark in which each example consists of $1$--$7$ premises and a conclusion, with a 3-way label (True / False / Uncertain) at $\approx 35{:}30{:}34\%$. We use the official validation split in full ($n{=}203$) and the same Qwen2.5-VL-3B-Instruct (\(a_{\mathrm{direct}}\), \(a_{\mathrm{retrieve}}\)) and Qwen2.5-VL-7B-Instruct (\(a_{\mathrm{defer}}\)) backbones in text-only mode, with rule-based per-action risk ($h{=}0.5$ if action's answer is wrong, $h{=}0$ otherwise). The 12-d feature block is per-action confidence margin, sequence log-prob, recover-context margin, three pairwise agreements, and two indicator features for the model's predicted-label class.

The full-$n$ quantities are $\alpha_{\mathrm{emp}}{=}0.687$, $\alpha_{\min}{=}0.633$, so $\beta\approx 0.0528$ (displayed as $0.053$ to three decimals throughout) and $n_{\min}^{\mathrm{Bern}}{=}1898$ from the leading-order formula at the unrounded $\beta$. FOLIO is therefore $\approx 9.4\times$ below the Bernstein threshold: Theorem~\ref{thm:bernstein} does \emph{not} certify the sign of the empirical \pit loss difference, and Corollary~\ref{cor:nested} item~(ii) predicts that whenever $\max_g p_g\gamma_g>0$ a partition router strictly improves on \piz while \pit may have wrong sign. Both predictions hold empirically: the deployable \pio winner KMeans-$K{=}6$ reaches $0.7195\pm 0.007$ versus \piz's $0.7520$ (a $0.0325$ reduction, $4.5$ seed-sd of \pio); the deployable \pit winner Selective-$C{=}0.3$ reaches $0.7546\pm 0.009$, $+0.0027$ \emph{worse} than \piz, exactly the uncertified-sign behaviour Corollary~\ref{cor:nested}~(ii) predicts when $n\ll n_{\min}^{\mathrm{Bern}}$.

The partition's anatomy is unbalanced and concentrated. The KMeans clustering recovers eight clusters at $K{=}8$; only three carry positive discriminability (the three clusters whose cell-best action is not the global best fixed action), with $p_g\gamma_g$ contributions $0.006,0.012,0.047$ summing to $\sum_g p_g\gamma_g{=}0.065$. The empirical KM-$K{=}6$ loss reduction $0.0325$ is below this oracle ceiling because (i) strict-CV refits KMeans on each outer-train fold of size $\sim 162$, smaller than the full-pool seed-$42$ partition, and (ii) the ceiling assumes oracle cell-best actions whereas the empirical estimator must learn them. \pit's wrong-sign behaviour is structurally driven: with $\beta{=}0.053$ the canonical \pit asymptotic ceiling is only $q^\star\beta(L_w{-}L_r)\approx 0.024$, which by Theorem~\ref{thm:bernstein} requires $n\ge 1898$ to be reliably recovered; at $n{=}203$ the variance overhead exceeds the ceiling. Theorem-driven prediction (\pio winner; \pit uncertified) and empirical realization match without further tuning.

\section{Controlled synthetic validation}
\label{app:synth}

We complement the per-benchmark results with three controlled synthetics that hold all but one structural variable fixed and sweep only the relevant diagnostic: Section~\ref{app:bernstein-cross} sweeps \(n\) across the Bernstein viability threshold of Theorem~\ref{thm:bernstein}; Section~\ref{app:synth-pi12} varies instance-level signal at fixed partition gap to test the \pio/\pit phase transition (Cor~\ref{cor:nested}~(iii)); Section~\ref{app:pi3-synth} varies an external prior channel's informativeness to characterize when \pith wins.

\subsection{Bottom-\texorpdfstring{$q$}{q} precision-estimator cross-threshold for Theorem~\ref{thm:bernstein}}
\label{app:bernstein-cross}

The four core benchmarks place a single point on each side of the Bernstein threshold (FOLIO below, HallusionBench/A-OKVQA above), but they cannot show how sign correctness behaves \emph{as} \(n\) crosses \(n_{\min}\). The synthetic in Section~\ref{app:synth-pi12} sweeps \pio/\pit instead and stays above threshold throughout. We therefore include a dedicated cross-threshold experiment on the same bottom-\(q\) precision estimator that Theorem~\ref{thm:bernstein} analyses (not the full \pit/CV pipeline; see Appendix~\ref{app:synth-pi12} for that).

At fixed \((\alpha, q, \delta)=(0.75, 0.3, 0.05)\), we sweep \(n\in\{20, 30, 50, 80, 130, 200, 320, 500, 800, 1300, 2000, 3200, 5000, 8000\}\) and \(\beta\in\{0.05, 0.10, 0.20\}\) (so \(n_{\min}\in\{1844, 461, 115\}\)). For each \((n, \beta)\), we draw \(4000\) replications of the bottom-\(q\) precision estimator (each replication: \(m=\lfloor nq\rfloor\) i.i.d.\ Bernoulli\((\alpha)\) draws under the local tail-margin convention) and record the empirical sign-correctness rate \(\Pr[\,\mathrm{sign}(\widehat{\Delta}(q))=\mathrm{sign}(\Delta(q))\,]\). Theorem~\ref{thm:bernstein} predicts this rate is at least \(1-\delta=0.95\) once \(n\ge n_{\min}(\beta)\).

\begin{figure}[t]
  \centering
  \includegraphics[width=0.6\linewidth]{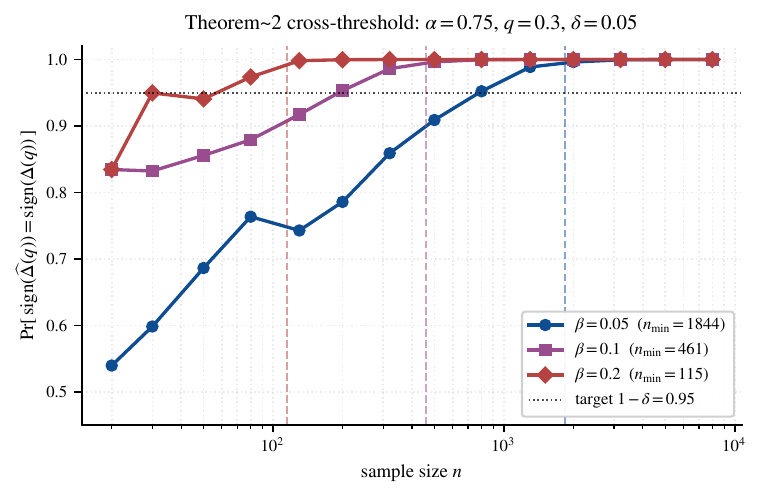}
  \caption{Bottom-\(q\) precision-estimator cross-threshold. At each \(\beta\), the empirical sign-correctness rate (over \(4000\) replications per cell) reaches the target \(1-\delta=0.95\) (dotted black) at sample sizes consistent with, and somewhat below, the predicted threshold \(n_{\min}(\beta)\) (dashed vertical lines). The empirical curves are consistent with the order-tight \(\beta^{-2}\) scaling guaranteed by Proposition~\ref{prop:lower} but show the constant slack typical of Bernstein-style sufficient conditions; the synthetic does not run the full \pit/CV pipeline and so does not directly verify constant-tightness in deployed controllers. Above each \(n_{\min}\) the rate saturates at \(\approx 1\); below, it degrades smoothly as predicted by the variance-bounded sub-case of Corollary~\ref{cor:nested}~(ii).}
  \label{fig:bernstein-cross}
\end{figure}

Figure~\ref{fig:bernstein-cross} shows the resulting rates. For each \(\beta\), the empirical curve crosses the \(1-\delta\) target at a sample size at or somewhat below \(n_{\min}(\beta)\), confirming the qualitative prediction (sign certified for \(n\gtrsim n_{\min}\), uncertified below) and showing scaling consistent with order-tight \(\beta^{-2}\) separation. The displayed grid's first empirical crossings sit at \(n/n_{\min}\in[0.43, 0.69]\), the constant-factor gap typical between Bernstein-derived sufficient conditions and the actual finite-sample saturation point; Proposition~\ref{prop:lower} rules out improvement in the leading order, not in constants. FOLIO's empirical wrong-sign \pit lands in the curve's variance-bounded regime: its \(n/n_{\min}{=}203/1898{=}0.107\) ratio sits below all three empirical-saturation crossings, so the wrong-sign observation is robust to the constant slack of the Bernstein sufficient condition. The synthetic uses \(\alpha{=}0.75\) versus FOLIO's \(\alpha_{\mathrm{emp}}{=}0.687\), so the mapping is qualitative; HallusionBench at \(n{=}920\), \(\beta{=}0.469\) sits well past saturation.

\subsection{\texorpdfstring{\pio/\pit}{Pi\_1/Pi\_2} phase transition for Corollary~\ref{cor:nested}}
\label{app:synth-pi12}

\begin{figure}[t]
  \centering
  \includegraphics[width=0.85\linewidth]{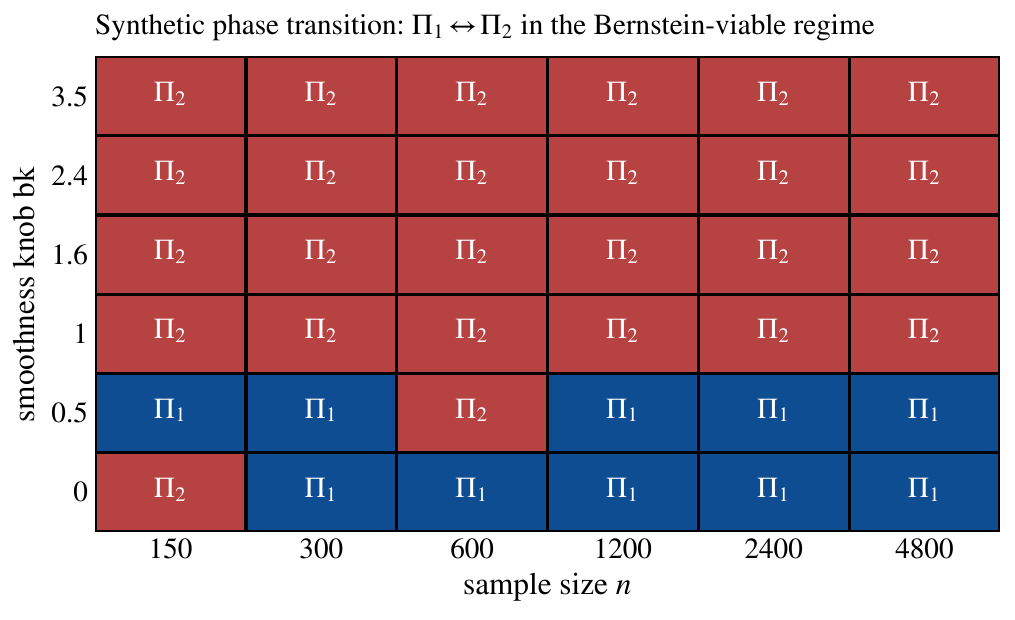}
  \caption{Controlled synthetic validation of Corollary~\ref{cor:nested} item~(iii). Each tile is one \((n,\text{bk})\) cell colored by the empirical winner class (blue \(=\!\pio\), red \(=\!\pit\)). The partition gap is held fixed at \(\max_g p_g\gamma_g\approx 0.08\); the smoothness knob bk on the y-axis is the only manipulated variable, and the resulting empirical margin \(\beta\) stays in \([0.37,0.62]\) across all cells, well above the Bernstein viability threshold of Theorem~\ref{thm:bernstein}, so this experiment is \emph{not} a Theorem~\ref{thm:bernstein} validation. At low bk the only signal is cluster identity and \pio's direct per-cluster argmin beats \pit's regularized LR softmax; at higher bk the added smooth feature gives \pit enough instance-level margin to overtake the partition gap. The empirical winner boundary thus realizes the direct comparison between instance-level margin and partition gain that Corollary~\ref{cor:nested} item~(iii) predicts.}
  \label{fig:synth}
\end{figure}

The data-generating process has four latent clusters of mass \((0.35,0.25,0.20,0.20)\) in \(\mathbb{R}^{6}\) with a residual heterogeneity that gives \pio a fixed population partition gap of \(\max_g p_g\gamma_g\approx 0.08\). A scalar knob \(\text{bk}\in\{0,0.5,1.0,1.6,2.4,3.5\}\) controls how much smooth continuous signal the features carry about \(c_{\mathrm{direct}}\): at \(\text{bk}=0\) the only information is cluster identity, and as bk grows a continuous feature is added that makes the discriminative signal increasingly extractable by a regularized LR classifier. We sweep \(n\in\{150,300,600,1200,2400,4800\}\) crossed with bk, run three seeds per cell with \pio realized as KMeans-\(K{=}4\) and \pit realized as a regularized LR softmax (a deliberately compact subset of the main-paper \pit pool, isolating capacity-realization effects), and declare the winner class. Every cell in the sweep sits comfortably above the Bernstein viability threshold of Theorem~\ref{thm:bernstein} (since \(\beta\) stays in \([0.37,0.62]\) throughout), so this experiment is \emph{not} a Theorem~\ref{thm:bernstein} validation. It is a direct test of the population-versus-empirical clause in item~(iii) of Corollary~\ref{cor:nested}: when the asymptotic ceilings $C_{\Pi_2}>C_{\Pi_1}$ at every cell, does the empirical winner still track the comparison once a finite-capacity \pit family is plugged in? Figure~\ref{fig:synth} plots the resulting winner map.

The result is precisely the population-versus-empirical gap that the second clause of Corollary~\ref{cor:nested}~(iii) anticipates. The asymptotic instance-level ceiling \(C_{\Pi_2}=q^\star\beta(L_w{-}L_r)\) exceeds the partition gap \(0.08\) at every cell since \(\beta\in[0.37,0.62]\) is large throughout, so the population-optimal clause of Cor~\ref{cor:nested}~(iii) predicts \pit. The empirical winner does not always match: at \(\text{bk}=0\), when the only source of margin is discrete cluster identity, \pio beats \pit at every \(n\ge 300\) because \pio's per-cluster argmin recovers the cell-best action sharply while the regularized LR family under-realizes the same discrete signal. As bk grows the LR family's recoverable margin rises (closer to its asymptotic ceiling), and once it overtakes the fixed partition gain the empirical winner flips to \pit and stays there: \pio wins \(10/12\) cells for \(\text{bk}\in\{0, 0.5\}\) and \pit wins all \(24/24\) cells for \(\text{bk}\ge 1\), with the two \pit wins at \((n{=}150, \text{bk}{=}0)\) and \((n{=}600, \text{bk}{=}0.5)\) within \(0.003\) of \pio (multi-seed noise). The synthetic therefore confirms that the empirical winner tracks the asymptotic-ceiling comparison only to the extent that the deployed \pit family realizes its ceiling; replacing the LR family with a higher-capacity \pit (HGBC), as in the main-paper pool, would shift the bk threshold but not the qualitative conclusion that ceiling comparison is mediated by family realization.

\subsection{\texorpdfstring{\pith}{Pi\_3} orthogonal-channel sweep}
\label{app:pi3-synth}

\pith can strictly improve on lower classes only when a label-free external prior channel carries information about correctness beyond what the \pit feature block provides. To characterize this condition directly, we sweep the prior's informativeness in a controlled DGP.

Inputs are drawn from the same four-cluster Gaussian as in Appendix~\ref{app:synth-pi12}, with the partition weakened (\(\mathrm{bump}{=}(+0.3,+0.3,+0.3,-0.3)\)) and the \pit smooth-signal knob fixed at \(\mathrm{bk}{=}1.6\), so \pit sits comfortably in its high-signal regime. We add a hidden prior scalar \(z\sim\mathcal{N}(0,1)\) drawn independently of \(X\), with correctness logit
\(\mathrm{logit}(c_{\mathrm{direct}}{=}1\mid X, z)=\mathrm{bk}\,s(X)+z_{\mathrm{strength}}\cdot z+\mathrm{bump}(g)\),
where \(s(X)\) is the unit-norm linear signal from Appendix~\ref{app:synth-pi12}. The feature block exposed to \piz, \pio, \pit is \(X\) only; \(z\) is exposed exclusively to \pith. Because \(z\) is independent of \(X\), no amount of \pit training on \(X\) can recover the information \(z\) carries about correctness, regardless of \(n\).

\pith is implemented as the Definition~\ref{def:classes} prior-gated controller with an explicit high-confidence gate: \(\pith(x,z){=}a_{\mathrm{direct}}\) if \(z>\tau\), \(\pith(x,z){=}a_{\mathrm{defer}}\) if \(z<-\tau\), and \(\pith(x,z){=}r(x)\) otherwise, with \(\tau{=}1.0\) so that the gate fires on the top/bottom \(\approx 16\%\) of the prior and leaves the middle \(\approx 68\%\) to the fallback. The asymmetric \(\mathrm{bump}=(+0.3,+0.3,+0.3,-0.3)\) keeps the partition discriminable so that the synthetic does not become trivially \pith-dominated by collapsing the \pio ceiling to zero. The fallback controller \(r \in \pit\) is a regularized multinomial logistic regression fit only on training samples with \(|z_{\mathrm{tr}}|\le\tau\). We cross \(n\in\{300,600,1200,2400,4800\}\) with \(z_{\mathrm{strength}}\in\{0,0.5,1.0,1.5,2.5,4.0\}\), 3 seeds per cell, under the same strict 5-fold CV protocol.

\begin{figure}[t]
  \centering
  \includegraphics[width=\linewidth]{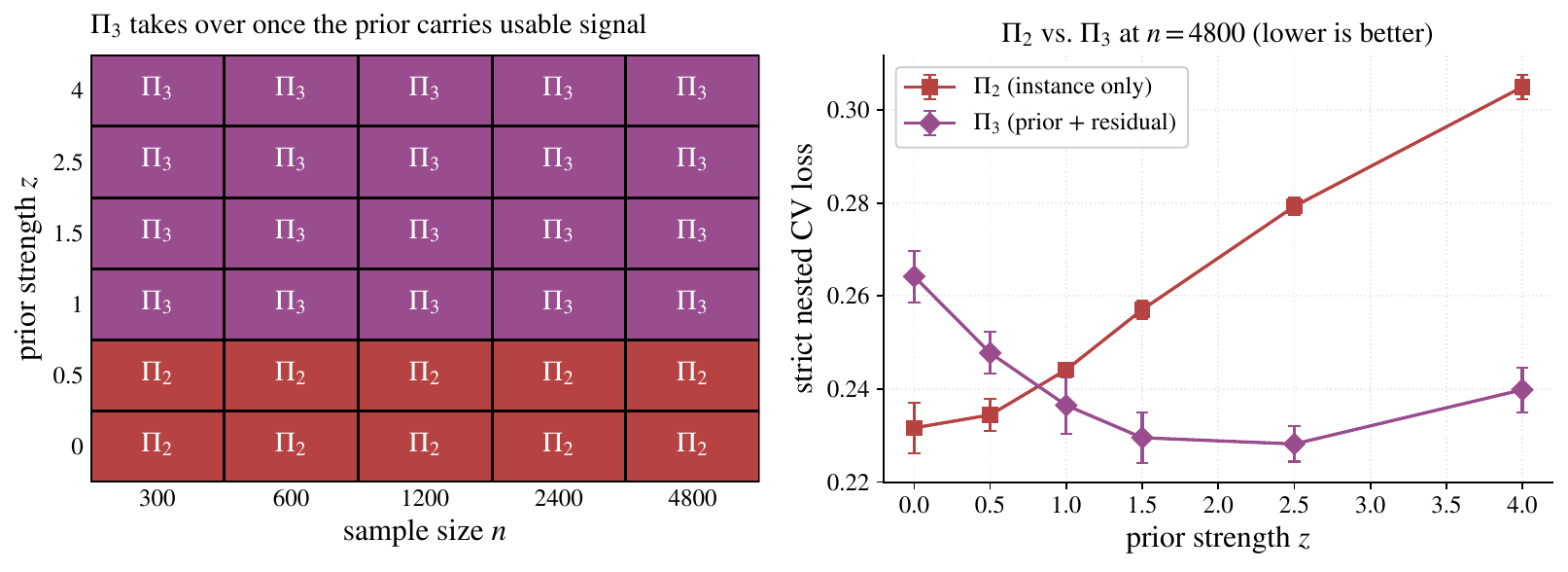}
  \caption{Controlled synthetic validation of \pith, with the partition signal weakened (\(|\mathrm{bump}|\le 0.3\)) and the \pit smooth-signal knob fixed at \(\mathrm{bk}{=}1.6\) so \pit sits comfortably in its high-signal regime. \textbf{Left:} empirical winner tile plot across \((n, z)\); each cell is a strict nested 5-fold CV outcome colored by the winning class (red\(=\!\pit\), purple\(=\!\pith\)). At \(z\in\{0,0.5\}\) the prior is uncorrelated or only weakly correlated with correctness and \pit wins every cell (10/30); at \(z\ge 1.0\) the prior carries enough signal about correctness that prior-gate decisions are more reliable than anything \pit can extract from \(X\), and \pith wins every cell (20/30). \textbf{Right:} at \(n{=}4800\), the \pit loss \emph{grows} from \(0.23\) to \(0.30\) as \(z\) increases (the correctness logit becomes increasingly \(z\)-dominated and \(X\) correspondingly less informative), while the \pith loss falls from \(0.26\) at \(z{=}0\) to the \(0.23\)--\(0.24\) band once \(z\ge1.0\). The separation between the two curves is the \pith gain.}
  \label{fig:pi3-synth}
\end{figure}

The winner map (Figure~\ref{fig:pi3-synth} left) partitions the sweep into two regimes: when \(z_{\mathrm{strength}}\le 0.5\) the prior is too weak to justify the gating cost and \pit wins uniformly in \(n\); when \(z_{\mathrm{strength}}\ge 1.0\) the prior is accurate enough that \pith's gated decisions dominate and \pith wins uniformly in \(n\). The transition is crisp: there is no ``\pith wins at large \(n\) but not small \(n\)'' axis, because \pith's advantage comes from additional information rather than from more samples. The right panel anchors the mechanism on \pith: at \(n{=}4800\), \pith's loss falls from \(0.264\) at \(z_{\mathrm{strength}}{=}0\) to the \(0.23\)--\(0.24\) band once \(z_{\mathrm{strength}}\ge1.0\), showing that the gate becomes useful only after the prior carries enough signal; \pit's loss rises from \(0.23\) to \(0.30\) by construction of the DGP, since fixing \(\mathrm{bk}{=}1.6\) while raising \(z_{\mathrm{strength}}\) makes \(X\) a relatively weaker driver of correctness, so the rising curve is a control rather than a finding. The displayed curves cross between \(z_{\mathrm{strength}}=0.5\) and \(1.0\) (linear interpolation at \(n{=}4800\) gives \(z_{\mathrm{strength}}\approx0.82\)), the operating point at which the prior carries enough signal that the certainty-on-the-tails gate outweighs the cost of routing the un-gated middle band to the fallback. The orthogonality \(z\perp X\) is the cleanest possible setting for \pith and overstates the advantage available to a real channel that shares partial information with \(X\); the TextVQA-OCR experiment in Section~\ref{sec:experiments} is the deployable partial-overlap test, and the A-OKVQA gold-rationale oracle in Appendix~\ref{app:pi3-rationale} is numerically consistent but methodologically non-deployable because the channel is answer-derived.

\section{Baselines, ablations, and CV diagnostics}
\label{app:robustness}

\subsection{Cost-sensitive learning-to-defer adaptations}
\label{app:l2d-results}

As a robustness check, we instantiate families from outside our own pool and ask whether they show the same qualitative failures and successes. We run two cost-sensitive adaptations of canonical learning-to-defer baselines under the identical strict nested CV protocol: a softmax surrogate adapted from Mozannar and Sontag \citep{mozannar2020consistent}, and a post-hoc plug-in estimator adapted from Narasimhan et al.\ \citep{narasimhan2022posthoc}. Both sit in \pit (instance-level learned controllers) by construction. The Mozannar variant is implemented as a row-replicated multinomial logistic regression with per-row weights \(w_{i,a}=L_{\max,i}-L[i,a]\); the Narasimhan variant trains one HistGradientBoosting regressor per action \(\widehat{L}_a(x)\) and selects \(\pi(x)=\arg\min_a\widehat{L}_a(x)\) at test time. These are not drop-in reproductions of the original human-deferral objectives; they are loss-matrix-compatible implementations designed to test whether other \pit-style learners encounter the same regimes. We sweep a small hyperparameter grid on each (\(C\in\{0.3,1.0\}\) for Mozannar, \texttt{max\_depth}\(\in\{3,4\}\) for Narasimhan) and run them through the same 5-fold-by-5-seed strict CV pipeline as our own families. Table~\ref{tab:l2d} reports the results.

\begin{table}[t]
  \centering
  \small
  \resizebox{\linewidth}{!}{
  \begin{tabular}{lcccc}
    \toprule
    Benchmark & Mozannar (best) & Narasimhan (best) & Best (ours) & Verdict \\
    \midrule
    HallusionBench & \(0.9203\pm0.005\) & \(0.8990\pm0.007\) & \(\mathbf{0.8970\pm0.004}\) (\pit) & Narasimhan ties HGBC ($0.3$ sd); Mozannar trails ($3.6$ sd) \\
    A-OKVQA        & \(0.3877\pm0.004\) & \(0.3756\pm0.002\) & \(\mathbf{0.3805\pm0.003}\) (\pit) & Narasimhan lower by $1.4$ sd; intra-\pit \\
    FOLIO          & \(0.7478\pm0.008\) & \(0.7705\pm0.021\) & \(\mathbf{0.7195\pm0.007}\) (\pio) & ours wins ($2.7$/$2.3$ joint sd) \\
    \bottomrule
  \end{tabular}}
  \caption{Cost-sensitive adaptations of two learning-to-defer baselines (Mozannar--Sontag 2020, Narasimhan et al.\ 2022) evaluated under the identical strict nested 5-fold-by-5-seed CV protocol used in this paper. Numbers are mean loss \(\pm\) seed std; lower is better. Bold marks our deployable per-benchmark winner from Table~\ref{tab:main}. Both adapted L2D baselines sit firmly in \pit by construction, so the comparison is intra-\pit on HallusionBench/A-OKVQA and inter-class on FOLIO. They recover the \pit-class result on HallusionBench (Narasimhan ties our HGBC within $0.3$ joint seed-sd; Mozannar trails by $3.6$ joint seed-sd) and on A-OKVQA (Narasimhan numerically below our Selective by $1.4$ joint seed-sd, intra-\pit variation), but trail our \pio winner on FOLIO by $2.7$ and $2.3$ joint seed-sd respectively, the variance-bounded sub-case (Cor~\ref{cor:nested}~(ii)) where instance-level estimation is unreliable and partition routing is the regime-correct class. SMS-Spam is excluded because in its residual-bounded regime (Theorem~\ref{thm:residual}) L2D wrappers reduce to direct by construction.}
  \label{tab:l2d}
\end{table}

The comparison is useful but deliberately limited. On HallusionBench ($n{=}920$, Bernstein-viable), Narasimhan-\texttt{md3} ($0.8990\pm 0.007$) ties our HGBC-\texttt{md3} ($0.8970\pm 0.004$) within $0.3$ joint seed-sd, while Mozannar-$C{=}0.3$ ($0.9203\pm 0.005$) trails by $\approx 3.6$ joint seed-sd; the spread reflects implementation-level differences (post-hoc plug-in vs.\ row-replication) within a single \pit class, not a class-selection signal. Our \pio KMeans-$K{=}8$ ($0.9093\pm 0.003$) beats Mozannar by $\approx 1.9$ joint seed-sd but trails Narasimhan within $\approx 1.4$ joint seed-sd, consistent with HallusionBench sitting just inside the high-signal \pit-favorable region. On A-OKVQA (high-signal boundary), Narasimhan-\texttt{md3} reaches $0.3756\pm 0.002$, numerically below our \pit Selective at $0.3805\pm 0.003$ by $\approx 1.4$ joint seed-sd. Both controllers are firmly \pit-class by construction, so this is intra-\pit variation rather than a class-selection signal: the canonical-pool \pit winner Selective and the external \pit implementation Narasimhan agree to within $0.005$, exactly the empirical-tightness pattern Corollary~\ref{cor:nested}~(iii) predicts when $C_{\Pi_2}$ is the ceiling and reasonable \pit families realize it; Mozannar at $0.3877$ shows that not every \pit implementation realizes the ceiling equally, an intra-\pit spread that does not flip the class-selection conclusion. On FOLIO (variance-bounded, $n{=}203\ll n_{\min}{=}1898$), Mozannar-$C{=}1.0$ ($0.7478\pm 0.008$) and Narasimhan-\texttt{md3} ($0.7705\pm 0.021$) trail our KMeans-$K{=}6$ partition at $0.7195\pm 0.007$ by $\approx 2.7$ and $\approx 2.3$ joint seed-sd respectively, matching the \pio-wins-\pit sub-case of Corollary~\ref{cor:nested}~(ii); both L2D adaptations are \pit-class by construction and inherit the $n<n_{\min}^{\mathrm{Bern}}$ uncertified-sign behaviour the theory predicts. The pattern across the three benchmarks tracks the theory: when $n\ge n_{\min}^{\mathrm{Bern}}$, multiple independent \pit implementations cluster within seed noise; when $n\ll n_{\min}$, partition routers strictly dominate the \pit realizations we tested, including the two L2D adaptations outside our main family pool. The adapted L2D baselines are not a proof that the original L2D objectives would behave identically under every implementation choice; they show that the observed regime classification is robust beyond our HGBC family.

\subsection{Ablations: CART partition router, loss-weight sensitivity}
\label{app:ablations}

Two ablations probe the robustness of the regime map. The first asks whether the \pio winners depend on KMeans. We add a second, structurally different \pio realization: a shallow CART partition router (\texttt{DecisionTreeClassifier} at \texttt{max\_depth}\(\in\{3,4\}\)) whose splits are fit on the per-sample loss-argmin label and whose leaves are re-assigned the cell-wise loss-argmin action at fit time (Appendix~\ref{app:tree}). Table~\ref{tab:tree} reports the per-benchmark numbers. On HallusionBench, KMeans-$K{=}8$ ($0.9093$) and CART-d4 ($0.9104$) tie within seed noise, both losing to the per-class \pit winner, so HallusionBench remains a clean \pit win regardless of partition method. On A-OKVQA the picture is more interesting: CART-d4 reaches $0.3742\pm 0.005$, numerically below the canonical \pit Selective at $0.3805\pm 0.003$ by $\approx 1.1$ joint seed-sd. This is within seed noise, so we retain Selective as the canonical \pit winner because the canonical pool is fixed in advance, but the CART ablation places A-OKVQA's deployable-class outcome in a \pio/\pit tie rather than a clean \pit win, sharpening the body's reading of A-OKVQA as the Cor~\ref{cor:nested}~(iii) ceiling-comparison boundary. It is exactly the second clause of the corollary at work: when $C_{\Pi_1}$ and $C_{\Pi_2}$ are commensurate, finite-capacity family realization---here the choice between an axis-aligned CART and a centroid-based KMeans---can flip the empirical winner between supportable adaptive classes. On FOLIO the depth-bounded CART overfits at $n{=}203$ and lands at $0.7981\pm 0.042$, $+0.046$ \emph{worse} than \piz; KMeans-$K{=}6$ at $0.7195$ remains the \pio winner. CART's FOLIO failure is the partition-router-side analogue of Cor~\ref{cor:nested}~(ii)'s \pit failure: at $n{=}203$, depth-3-or-4 CART leaves carry too few samples for the empirical per-leaf argmin to track its cell-best action, consistent with the per-leaf sample budget required by Theorem~\ref{thm:partition} not being met; KMeans's fixed $K$ is an implicit regularizer that the depth-bounded tree, with its more flexible leaf structure, lacks.

\begin{table}[t]
  \centering
  \small
  \resizebox{\linewidth}{!}{
  \begin{tabular}{lcccc}
    \toprule
    Benchmark & \piz best & KMeans \pio best & CART \pio best & verdict \\
    \midrule
    HallusionBench & \(1.0000\) & \(0.9093\pm0.003\) & \(0.9104\pm0.004\) & both beat \piz; tied within noise\\
    A-OKVQA        & \(0.4138\) & \(0.3902\pm0.009\) & \(\mathbf{0.3742\pm0.005}\) & both beat \piz; CART better\\
    FOLIO          & \(0.7520\) & \(\mathbf{0.7195\pm0.007}\) & \(0.7981\pm0.042\) & KMeans wins; CART overfits\\
    \bottomrule
  \end{tabular}}
  \caption{Non-KMeans \pio realization: shallow CART partition routers under the identical strict nested 5-fold-by-5-seed CV protocol. Bold marks the \pio winner among KMeans/CART. On HallusionBench KMeans and CART tie within seed noise. On A-OKVQA CART-d4 at $0.3742$ falls $\approx 1.1$ joint seed-sd below the canonical \pit winner Selective at $0.3805$, an additional witness for the Cor~\ref{cor:nested}~(iii) commensurate-ceiling boundary in which finite-capacity family choice can flip the empirical winner. On FOLIO, the depth-bounded tree overfits at $n{=}203$, consistent with Theorem~\ref{thm:partition}'s required per-leaf sample budget not being met; KMeans's fixed $K$ acts as an implicit regularizer.}
  \label{tab:tree}
\end{table}

The second ablation addresses the cost-sensitivity of the combined loss. Classical reject-option and cost-sensitive evaluation explicitly depend on operating costs \citep{chow2003optimum, drummond2006costcurves}, so a class-selection result should not hinge on a single arbitrary weight vector. We therefore rerun the deployable \(\piz/\pio/\pit\) family pool under seven local perturbations of the canonical core weights: \((1,1,0.05)\), correctness weight \(0.75\) and \(1.25\), risk weight \(0.75\) and \(1.25\), and cost weight \(0.025\) and \(0.10\), always with the same strict 5-fold-by-5-seed protocol. Table~\ref{tab:weight-sensitivity} summarizes the winner counts. Two of the three benchmarks are fully stable: A-OKVQA \pit ($7/7$) and FOLIO \pio ($7/7$). HallusionBench is the boundary case: the canonical-weights winner is \pit ($5/7$), but two perturbations (\texttt{correct\_high}, \texttt{risk\_low}) swap the winner to \pio by less than $0.005$, reflecting the fact that the canonical-weights gap between \pio (KM-$K{=}8$ at $0.909$) and \pit (HGBC-\texttt{md3} at $0.897$) is only $\approx 3$ seed-sd of \pit. The qualitative regime map is therefore robust on A-OKVQA and FOLIO, with HallusionBench's \pio/\pit boundary slightly weight-dependent: the operating-cost neighborhood swept here, which spans $\pm 25\%$ perturbations of the correctness and risk weights and a $2\times$ range of cost weights around the canonical setting, leaves the deployable winner unchanged on A-OKVQA and FOLIO but flips it on HallusionBench in two of seven variants---a quantitative confirmation that HallusionBench sits on the \pio/\pit ceiling-comparison boundary predicted by Corollary~\ref{cor:nested}~(iii).

\begin{table}[t]
  \centering
  \small
  \resizebox{\linewidth}{!}{
  \begin{tabular}{lcccc}
    \toprule
    Benchmark & canonical class & winner counts over 7 weights & min winner margin & interpretation \\
    \midrule
    HallusionBench & \pit & \pit:5, \pio:2 & 0.0047 & \pio/\pit boundary\\
    A-OKVQA        & \pit & \pit:7          & 0.0049 & stable \pit\\
    FOLIO          & \pio & \pio:7          & 0.0074 & stable \pio\\
    \bottomrule
  \end{tabular}}
  \caption{Loss-weight sensitivity on the three benchmarks where adaptive controllers are theoretically viable (SMS-Spam excluded as residual-bounded). We rerun the deployable \(\piz/\pio/\pit\) family pool under strict 5-fold-by-5-seed CV for seven local perturbations of the canonical weights \((1,1,0.05)\). The HallusionBench \pio/\pit boundary is itself a quantitative diagnostic: under the canonical weights \pit edges \pio by only $\approx 3$ seed-sd, and two perturbations swap the winner. A-OKVQA and FOLIO are fully stable across the seven variants, with FOLIO's $0.0074$ minimum margin reflecting its $4.5$ seed-sd canonical \pio gap.}
  \label{tab:weight-sensitivity}
\end{table}

\subsection{Automated class selection}
\label{app:auto-select}

Automated inner-CV class selection is near-best on the two Bernstein-viable benchmarks and predictably fails on FOLIO without the regime-theoretic class restriction. Strict nested auto-pick loss differences relative to \texttt{always\_direct} are \(\Delta\in\{-0.183, -0.028, -0.003\}\) on \{HallusionBench, A-OKVQA, FOLIO\}. On HallusionBench the auto-pick lands within seed noise of the per-class \pit winner ($\Delta{=}-0.183$ vs.\ per-class \pit winner $\Delta{=}-0.188$, $0.7$ joint seed-sd), and A-OKVQA remains close but more borderline ($\Delta{=}-0.028$ vs.\ per-class \pit winner $\Delta{=}-0.033$, $1.4$ joint seed-sd). On FOLIO, the auto-pick reaches only $\Delta{=}-0.003$, materially below the per-class \pio winner at $\Delta{=}-0.033$ (a $0.030$ absolute-loss gap, or $1.7$ joint seed-sd when auto-pick variability is included) because at $n{=}203$ the unrestricted inner-CV pool occasionally lands on \pit families whose sign Theorem~\ref{thm:bernstein} explicitly does not certify. This is exactly the failure mode Corollary~\ref{cor:nested}~(ii) predicts when $n<n_{\min}^{\mathrm{Bern}}$: empirical CV alone cannot reliably separate \pio from \pit, and the regime diagnostic is what supplies the missing class restriction. The same per-fold class shares ($19/25$ \pio, $4/25$ \pit, $2/25$ \piz) make this concrete: $4$ outer-folds pick a \pit family at coverage~$q$ where Theorem~\ref{thm:bernstein} does not certify the sign, and these $4$ folds are what drag the auto-pick mean from the per-class \pio winner to the much smaller $-0.003$. Restricting the candidate pool with the regime diagnostic ($\beta>0,\, n<n_{\min}^{\mathrm{Bern}} \Rightarrow$ exclude \pit per Cor~\ref{cor:nested}~(ii)) would force those $4$ \pit folds to be reselected from \(\piz\cup\pio\), removing the uncertified instance-level choices and steering selection toward the regime-correct \pio-dominated pool. \emph{HallusionBench} picks $\pit$ HGBC on $20/25$ outer-folds (HGBC-md4 on $12/25$, HGBC-md3 on $8/25$) and $\pio$ KMeans on the remaining $5/25$, consistent with the per-class \pit winner in Table~\ref{tab:main}. \emph{FOLIO} picks $\pio$ KMeans on $19/25$ outer-folds (KM-$K{=}6$ on $6/25$, KM-$K{=}8$ on $5/25$, KM-$K{=}4$ and $K{=}5$ on $4/25$ each), the \pit family Selective-$C{=}0.3$ on $4/25$, and \piz \texttt{always\_direct} on $2/25$; this dominant \pio class share matches the deployable per-class winner. \emph{A-OKVQA} is the one benchmark where the inner-CV class selection is ambiguous: $10/25$ outer-folds pick the \pio winner KMeans-$K{=}4$ while the remaining $15/25$ are distributed across three \pit families (Selective $C{=}0.3$ on $10/25$, HGBC-md3 on $4/25$, HGBC-md4 on $1/25$), giving a $10{:}15$ \pio:\pit class split with the resulting auto-pick loss falling between the two per-class winners. This is precisely the high-signal \pio/\pit boundary case predicted by Corollary~\ref{cor:nested} item~(iii) when the \pio and \pit ceilings are commensurate; the per-class decomposition on outer-train remains a more informative summary than the inner-CV argmin in this regime. On TextVQA-OCR, the per-class decomposition selects \pith as the deployable top-rung winner because the OCR channel is available at prediction time. If the non-deployable A-OKVQA gold-rationale oracle is included in the candidate pool, inner-CV also selects it; we report that only as the oracle experiment of Appendix~\ref{app:pi3-rationale}, not as the deployable auto-pick result.

\end{document}